\documentclass[10pt,journal]{IEEEtran}
\usepackage{amsmath,amsfonts}
\usepackage{algorithmic}
\usepackage{algorithm}
\usepackage{setspace}
\usepackage{array}
\usepackage[caption=false,font=normalsize,labelfont=sf,textfont=sf]{subfig}
\usepackage{textcomp}
\usepackage{stfloats}
\usepackage{url}
\usepackage{verbatim}
\usepackage{graphicx}
\usepackage{tablefootnote}
\usepackage[flushleft]{threeparttable}
\usepackage{tabularx,booktabs} 
\usepackage[numbers]{natbib}
\usepackage{microtype}
\usepackage{scalerel}
\usepackage{tikz}
\usepackage{bm}
\usepackage{enumitem}
\usepackage{chngcntr}
\usepackage{float}
\usepackage[countmax]{subfloat}
\usepackage{multirow}
\usepackage{multicol}
\usepackage{xcolor}
\usepackage{hyperref}
\usepackage{pifont}
\newcommand{\cmark}{\ding{51}}%
\newcommand{\xmark}{\ding{55}}%

\usetikzlibrary{svg.path}
\definecolor{orcidlogocol}{HTML}{A6CE39}
\tikzset{
  orcidlogo/.pic={
    \fill[orcidlogocol] svg{M256,128c0,70.7-57.3,128-128,128C57.3,256,0,198.7,0,128C0,57.3,57.3,0,128,0C198.7,0,256,57.3,256,128z};
    \fill[white] svg{M86.3,186.2H70.9V79.1h15.4v48.4V186.2z}
                 svg{M108.9,79.1h41.6c39.6,0,57,28.3,57,53.6c0,27.5-21.5,53.6-56.8,53.6h-41.8V79.1z M124.3,172.4h24.5c34.9,0,42.9-26.5,42.9-39.7c0-21.5-13.7-39.7-43.7-39.7h-23.7V172.4z}
                 svg{M88.7,56.8c0,5.5-4.5,10.1-10.1,10.1c-5.6,0-10.1-4.6-10.1-10.1c0-5.6,4.5-10.1,10.1-10.1C84.2,46.7,88.7,51.3,88.7,56.8z};
  }
}
\newcommand\orcidicon[1]{\href{https://orcid.org/#1}{\mbox{\scalerel*{
\begin{tikzpicture}[yscale=-1,transform shape]
\pic{orcidlogo};
\end{tikzpicture}
}{|}}}}

\begin{document}
\markboth{IEEE INTERNET OF THINGS JOURNAL, ~Vol.~X, No.~X, X~2023}
{Shell \MakeLowercase{\textit{et al.}}: A Sample Article Using IEEEtran.cls for IEEE Journals}

\title{An Empirical Study of Federated Learning on IoT-Edge Devices: Resource Allocation and Heterogeneity}
\noindent\author{
Kok-Seng Wong$^{*}$ \orcidicon{0000-0002-2029-7644}, \textit{Member, IEEE}, Manh Nguyen-Duc \orcidicon{0000-0002-9668-4974}, Khiem Le-Huy \orcidicon{0000-0001-8625-6864}, Long Ho-Tuan \orcidicon{0009-0003-4948-5395}, \\Cuong Do-Danh \orcidicon{0000-0003-4785-4524}, \textit{Member, IEEE}, and Danh Le-Phuoc \orcidicon{0000-0003-2480-9261}, \textit{Member, IEEE}

\thanks{Manuscript received X 2023; revised X 2023; accepted X 2023. Date of publication X 2023; date of current version X 2023. 
}
\thanks{Kok-Seng Wong, Khiem Le-Huy, Long Ho-Tuan, and Cuong Do-Danh are with the College of Engineering and Computer Science, VinUniversity, Hanoi, Vietnam. (e-mail:\{wong.ks, khiem.lh, long.ht, cuong.dd\}@vinuni.edu.vn)}
\thanks{Manh Nguyen-Duc, and Danh Le-Phuoc are with the Technische Universität Berlin Straße des 17. Juni 135, 10623 Berlin, Germany. (e-mail: manh.nguyenduc@campus.tu-berlin.de, danh.lephuoc@tu-berlin.de)}
\thanks{Kok-Seng Wong, Manh Nguyen-Duc, and Khiem Le-Huy contributed equally to this work. }
\thanks{$*$ Corresponding author: wong.ks@vinuni.edu.vn (Kok-Seng Wong)}
}
\maketitle

\begin{abstract}
Nowadays, billions of phones, IoT and edge devices around the world generate data continuously, enabling many Machine Learning (ML)-based products and applications. However, due to increasing privacy concerns and regulations, these data tend to reside on devices (clients) instead of being centralized for performing traditional ML model training. Federated Learning (FL) is a distributed approach in which a single server and multiple clients collaboratively build an ML model without moving data away from clients. Whereas existing studies on FL have their own experimental evaluations, most experiments were conducted using a simulation setting or a small-scale testbed. This might limit the understanding of FL implementation in realistic environments. In this empirical study, we systematically conduct extensive experiments on a large network of IoT and edge devices (called IoT-Edge devices) to present FL real-world characteristics, including learning performance and operation (computation and communication) costs. Moreover, we mainly concentrate on heterogeneous scenarios, which is the most challenging issue of FL. By investigating the feasibility of on-device implementation, our study provides valuable insights for researchers and practitioners, promoting the practicality of FL and assisting in improving the current design of real FL systems. 
\end{abstract}

\begin{IEEEkeywords}
Federated Learning,  IoT-Edge Devices, On-Device Training, Empirical Study.
\end{IEEEkeywords}

\section{Introduction}
\label{sec:intro}
\IEEEPARstart{B}{y} the end of 2018, there were an estimated 22 billion IoT devices in use around the world and this number is increasing fast. Forecasts suggest that by 2030 the number of IoT devices will increase to around 50 billion ~\cite{iotdevices}. Also, 100 billion ARM CPUs currently dominate the IoT market have been shipped so far ~\cite{100barms}. This installation base is a key enabler for many industrial and societal domains, especially Artificial Intelligence (AI) and Machine Learning (ML) powered applications ~\cite{DBLP:journals/cacm/Groger21}. However, due to increasing privacy concerns and regulations ~\cite{GDPR}, especially in sensitive domains like healthcare or finance, these valuable assets mostly remain inaccessible and cannot be centralized for conducting traditional ML model training. 

\begin{figure}[t]
    \centering
    \includegraphics[keepaspectratio, width=0.5\textwidth]{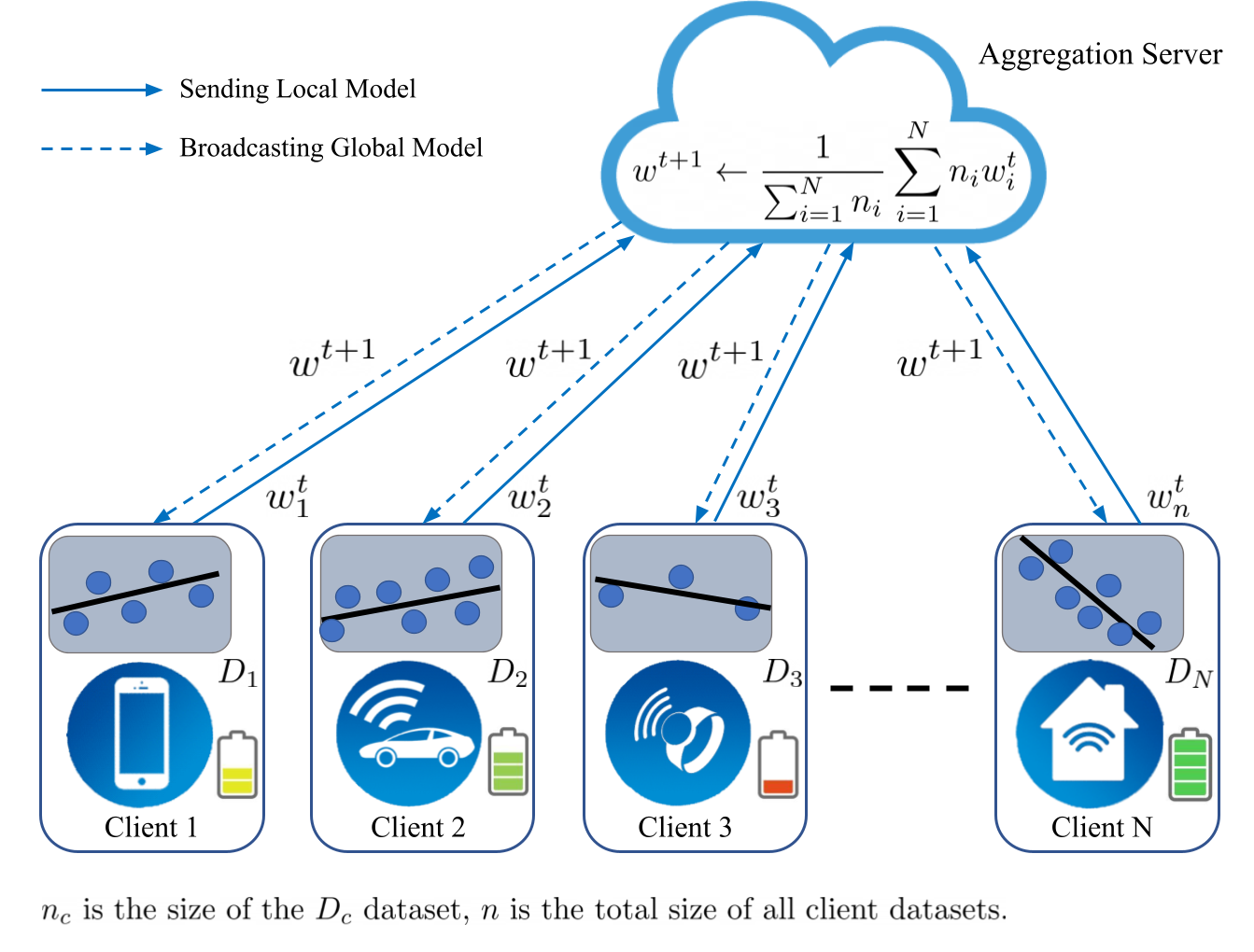}
    \caption{The standard FL framework. }
    \label{fig:fl}
\end{figure}

To address this issue, Federated Learning (FL) \cite{fedavg} was proposed, which allows multiple parties (clients) to train a shared global model collaboratively in a decentralized fashion without sharing any private dataset.  In general, a standard FL framework, as illustrated in Fig. \ref{fig:fl}, consists of two main steps: (1) Client training, in which clients train models on their local data for several epochs and send their trained models to a central server, and (2) Model aggregation, in which the server aggregates those models to establish a global model and distributes this global model back to the clients. This 2-step procedure is repeated for numerous rounds until the global model converges or a target level of accuracy is reached. 

Although FL recently has received considerable attention from the research community \cite{Survey-Concept, Survey-Vision} thanks to several advantages such as scalability or data privacy protection, it still has many serious challenges which lead to difficulties for real-world implementation. Specifically, clients in a federation differ from each other in terms of computational and communication capacity. For instance, the hardware resources (memory, CPU/GPU, or connectivity) of various IoT and edge devices (IoT-Edge devices) like Raspberry Pi devices or NVIDIA Jetson devices are much different. Therefore, considering all clients equally might lead to suboptimal efficiency. Furthermore, the training data owned by each client can be non-independent, identically distributed (Non-IID), and with different quality and quantity. These challenges make FL impractical and limit the motivation of parties to join the federation for training. 


Despite the aforementioned real-world issues, most existing studies on FL heavily rely on simulation settings or small-scale testbeds of devices \cite{fedadapt, fedmsa, ce-fedavg} to examine the behavior of their systems. While simulation settings are useful for controlled testing and development of FL models, they face significant challenges in adequately covering all operational aspects of real-world deployments. Specifically, existing simulators cannot emulate crucial aspects of realistic execution environments, such as resource consumption (e.g., memory, CPU/GPU usage, battery life) and network connectivity (e.g., bandwidth and network congestion). These factors significantly impact the performance of FL systems, as demonstrated in Section~\ref{sec:results}. Additionally, other realistic environment aspects such as data distribution, underlying software libraries, and executing settings introduce further challenges that can affect FL performance. Therefore, this motivates us to conduct more comprehensive evaluations of such aspects to ensure their effectiveness and scalability.

In Section~\ref{sec:related-works}, we observe a lack of experimental studies that systematically investigate the implementation of FL on real devices and assess the impact of intrinsic heterogeneity on performance and costs. Although there have been some attempts to implement FL on IoT-Edge devices at small scales with simplistic settings, it is desirable to have more reproducible experiments in larger and more realistic settings.  Hence, to the best of our knowledge, our study pushed the experiment scale and complexity to a new level.


\subsection{ Objectives, Research Questions and Scope}
\label{sec:objectives_RQ}

To identify potential issues and limitations on real devices that may not be apparent in simulated environments,  we focus our study on the impact of resource allocations and heterogeneity independently and their combined effects in realistic environments. To achieve this, we focus on the following research questions (RQ): 
\begin{itemize}[leftmargin=*]
    \item \textbf{RQ1: What are the behaviors of FL implementation in realistic environments compared to a simulation setting?} In this RQ, we compare many simulation and on-device deployment aspects. We want to see how simulation results can represent reality because FL experiments conducted in a controlled laboratory setting may not accurately reflect the challenges and complexities of realistic device-based environments. 
    \item \textbf{RQ2: How do resource allocation and heterogeneity affect the learning performance and operation costs?} There are several factors that can affect FL deployment. This RQ focuses on the client participation rate, communication bandwidth, device and data heterogeneity. We test each factor independently to learn their impact on the behaviors of FL. Specifically, we want to observe the impact of varying the number and type of devices, bandwidth, and data distribution on the FL process for each factor.
    \item \textbf{RQ3: How do these two factors, resource allocation and heterogeneity, simultaneously affect the learning performance and operation costs?} This RQ is an essential study on understanding the impact of combined factors as specified in RQ2. Additionally, we aim to find the dominant factor towards the behaviors of FL in a real-world deployment. 
\end{itemize}


To answer these questions, we need stable FL systems that can be deployed our targeted hardware, i.e., Raspberry Pi 3 (Pi3), Raspberry Pi 4 (Pi4), Jetson Nano (Nano) and Jetson TX2 (TX2) and can support GPUs on edge computing boards. While many algorithms are accompanied by source code, only Federated Averaging (FedAvg) \cite{fedavg} can satisfy our requirements due to its popularity.
FedAvg has been extensively studied and evaluated in the literature with a large number of works reporting its performance characteristics and limitations in simulations. However, understanding its behavior on real devices is still limited (c.f. Secion~\ref{sec:related-works}. Hence, we will focus on FedAvg for our studies in this paper and leave others for future work. However, our experiment design in Section~\ref{sec:exp-designs} is general enough to be replicated in other algorithms, given that their implementations are stable enough to run on targeted devices.

\subsection{Our Key Findings}
Along this light,  our extensive set of experiments reported in Section~\ref{sec:results} 
reveal the following key findings:
\begin{itemize}[leftmargin=*]
    \item The on-device settings can achieve similar training accuracy to the simulation counterparts with similar convergence behaviors.
    But when it comes to operational behaviours related to computation and communication, the on-device ones show much more complicated behavior patterns for realistic IoT-Edge deployments.
    \item The disparity in computational and networking resources among the participating devices leads to longer model update (local and global) exchange times because high computational devices need to wait for the server to receive and aggregate local updates from low computational devices. This hints that an oversimplified emulation of these aspects in simulation setting highly likely lead to unexpected outcomes of a FL algorithm at the deployment phase.
    \item Data heterogeneity is the most dominant factor in FL performance, followed by the number of clients. The performance of the global model is affected most by the data distribution (i.e., Non-IID and Extreme Non-IID) of each participating client, especially for challenging learning tasks. Hence, combining with the  disparity in computational and networking resources, FL on diverse IoT-Edge devices in realistic deployment settings need further understanding on-device behaviours in terms combining all these factors in tandem. 
\end{itemize}

\subsection{Paper Outline}
The rest of this article is organized as follows. Section \ref{sec:related-works} presents preliminaries to our work and discusses some existing surveys and empirical studies on FL. In Section \ref{sec:exp-designs}, we show our experimental designs and followed by our results and findings in Section \ref{sec:results}. Finally, we give further discussions in Section \ref{sec:discussions} and conclude this empirical study in Section \ref{sec:conclusion_limitation_futurework}. 

\section{Preliminaries and Related Works}
\label{sec:related-works}

\subsection{Federated Learning}
In the standard FL framework, data for learning tasks is acquired and processed locally at the IoT-Edge nodes, and only the trained model parameters are transmitted to the central server for aggregation. In general, along with an initialization stage, FL involves the following stages: 
\begin{itemize}[leftmargin=*]
    \item \textit{Stage 0 (Initialization}): The aggregation server $S$ first initiates the weight $w_0$ of the global model and hyperparameters such as the number of communication rounds $T$, size of the selected clients for each round $N$, and local training details. 
    \item \textit{Stage 1 (Client training}): All selected clients $C_1$, $C_2$, $C_3$, ..., $C_N$ receive the current global weight from $S$. Next, each $C_i$ updates its local model parameters $w_i^t$ using its local dataset $D_i$, where $t$ denotes the current communication round. Upon the completion of the local training, all selected clients send the local weight to $S$ for model aggregation. 
    \item \textit{Stage 2 (Model Aggregation)}: $S$ aggregates the received local weights based on a certain mechanism and then sends back the aggregated weights to the clients for the next round of local training. 
\end{itemize}

\subsection{Federated Averaging Algorithm}
Federated Averaging (FedAvg) is the de facto FL algorithm that is included in most FL systems \cite{fedavg}. As shown in Algorithm \ref{fedavg}, FedAvg aggregates the locally trained model parameters by weighted averaging proportional to the amount of local dataset $D_i$, that each client $C_i$ had (corresponding to the above Stage 2). Note that there are many advanced FL algorithms were introduced (e.g., FedProx \cite{fedprox} and FedMA \cite{fedma}), with different purposes in the last few years \cite{Chen-survey, Liu-survey}. 

\begin{algorithm}[!ht]
\setstretch{1.1}
\caption{FedAvg Algorithm \cite{fedavg}. }
\label{fedavg}
\begin{algorithmic}[1]
\STATE \textbf{Aggregation Server executes:}
\STATE initialize: $w \gets w_0$
\FOR{each round $t=1,2,3,\dots,T$}
\FOR{each client $i=1,2,3,\dots,N$ \textbf{in parallel}}
\STATE $w_i^t \gets w^{t-1}$
\STATE $w_i^t \gets$ \textbf{ClientTraining}($w_i^t$, $D_i$)
\ENDFOR
\STATE // ModelAggregation
\STATE $ w^{t+1} \gets \frac{1}{\sum_{i=1}^{N} n_i} \sum_{i=1}^{N} n_i w_i^t $ 
\ENDFOR
\STATE return: $w^T$
\STATE
\STATE \textbf{ClientTraining}($w_i$, $D_i$)\textbf{:} // Run on client $C_i$
\FOR{each epoch $e=1,2,3,\dots,E$}
\STATE $w_i \gets w_i - \eta \nabla l(w_i; D_i)$
\ENDFOR
\STATE return: $w_i$
\end{algorithmic}
\end{algorithm}

\subsection{Related Works}

Several available theoretical surveys and simulation-based empirical studies on FL are available in the literature. Dinh et al. \cite{fl-iot-1} explore and analyze the potential of FL for enabling a wide range of IoT services, including IoT data sharing, data offloading and caching, attack detection, localization, mobile crowdsensing, and IoT privacy and security. Ahmed et al. \cite{fl-iot-2} discuss the implementation challenges and issues when applying FL to an IoT environment. Zhu et al. \cite{noniid-survey} provides a detailed analysis of the influence of Non-IID data on different types of ML models in both horizontal and vertical FL. Li et al. \cite{noniid-study} conduct extensive experiments to evaluate state-of-the-art FL algorithms on Non-IID data silos and find that Non-IID does bring significant challenges in learning accuracy of FL algorithms, and none of the existing state-of-the-art FL algorithms outperforms others in all cases. Recently, Matsuda et al. \cite{personalized-study} benchmark the performance of existing personalized FL through comprehensive experiments to evaluate the characteristics of each method and find that there are no champion methods. Caldas et al. \cite{leaf} propose LEAF, a modular benchmarking simulation-based framework for learning in federated settings. LEAF includes a suite of open-source federated datasets, a rigorous evaluation framework, and a set of reference implementations. To the best of our knowledge, we are the first ones that consider an empirical study of FL on IoT-Edge devices. 

For real-world FL implementation, Di et al. \cite{fedadapt} present FedAdapt, an adaptive offloading FL framework based on reinforcement learning and clustering to identify which layers of the DNN should be offloaded for each device onto a server. Experiments are carried out on a lab-based testbed, including two Pi3s, two Pi4s, and one Jetson Xavier. Sun et al. \cite{fedmsa} propose a model selection and adaptation system for FL (FedMSA), which includes a hardware-aware model selection algorithm, then demonstrate the effectiveness of their method on a network of two Pi4s and five Nanos. Mills et al. \cite{ce-fedavg} propose adapting FedAvg to use a distributed form of Adam optimization, then test their method on a small testbed of five Pi2s and five Pi3s. Furthermore, Zhang et al. \cite{fliot} build the FedIoT platform for on-device anomaly data detection and evaluate their platform on a network of ten Pi4s. However, these attempts are still on a small scale and do not represent real-world environments. 


\begin{table}[!ht]
\caption{Comparison between our work and others. }
\setlength{\tabcolsep}{3.66pt}
\renewcommand{\arraystretch}{1.1}
\label{tab:related-work}
\begin{threeparttable}
\begin{tabular}{|l|c|ccc|}
\hline
\multirow{2}{*}{\begin{tabular}[c]{@{}l@{}}Empirical\\ Studies\end{tabular}} & \multirow{2}{*}{\begin{tabular}[c]{@{}c@{}}Simulation\\ -based\end{tabular}} & \multicolumn{3}{c|}{Device-based}                                                                                                                                                                                                              \\ \cline{3-5} 
                                                                             &                                                                              & \multicolumn{1}{c|}{\begin{tabular}[c]{@{}c@{}}Small-scale\\ (up to 10)\end{tabular}} & \multicolumn{1}{c|}{\begin{tabular}[c]{@{}c@{}}Large-scale\\ (up to 64)\end{tabular}} & \begin{tabular}[c]{@{}c@{}}Device\\Heter.\tnote{*}\end{tabular} \\ \hline
\cite{noniid-study, personalized-study, leaf} & \cmark                                                                        & \multicolumn{1}{c|}{\xmark}                                                            & \multicolumn{1}{c|}{\xmark}                                                            & \xmark                                                          \\ \hline
\cite{fliot, fedmsa, ce-fedavg}  & \cmark                                                                        & \multicolumn{1}{c|}{\cmark}                                                            & \multicolumn{1}{c|}{\xmark}                                                            & \xmark                                                          \\ \hline
Ours                                                                         & \cmark                                                                        & \multicolumn{1}{c|}{\cmark}                                                            & \multicolumn{1}{c|}{\cmark}                                                            & \cmark                                                          \\ \hline
\end{tabular}
\begin{tablenotes}
    \item[*] Device Heterogeneity: Study of different types of IoT devices
\end{tablenotes}
\end{threeparttable}
\end{table}

\section{Experimental Design}
\label{sec:exp-designs}

\begin{figure*}[!ht]
    \centering
    \includegraphics[keepaspectratio, width=1.0\textwidth]{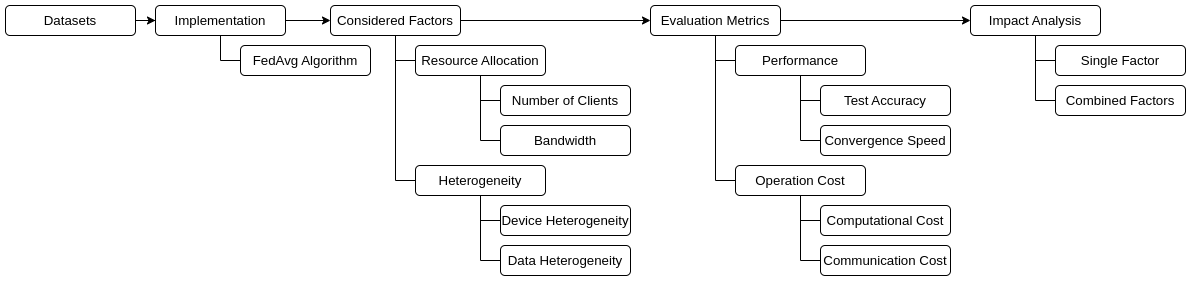}
    \caption{Our Methodology. }
    \label{fig:metho}
\end{figure*}

This section describes how we designed our experiments to answer our research questions in Section~\ref{sec:objectives_RQ}. Starting with data preparation, we then implement FL on IoT-Edge devices with different settings based on the  evaluation factors we defined. After that, we use a bag of metrics to analyze the impact of these factors individually and their combined effects in different aspects. Fig. \ref{fig:metho} illustrates this workflow in detail. 

\subsection{Data Preparation and Models}

\subsubsection{Datasets}
We use two datasets in this work 
: CIFAR10 \cite{cifar} and CIFAR100 \cite{cifar}, which are commonly used in previous studies on FL \cite{fedprox, noniid-study}. CIFAR10 consists of 60000 32x32 color images and is the simple one. The images are labeled with one of 10 exclusive classes. There are 6000 images per class with 5000 training and 1000 testing images. CIFAR100 also consists of 60000 32x32 color images and is more challenging to train, however, each image comes with one of 100 fine-grained labels. There are 600 images per class with 500 training and 100 testing images. 

\subsubsection{Data Partitioning}
\begin{figure*}[!ht]
    \centering
    \includegraphics[keepaspectratio, width=0.99\textwidth]{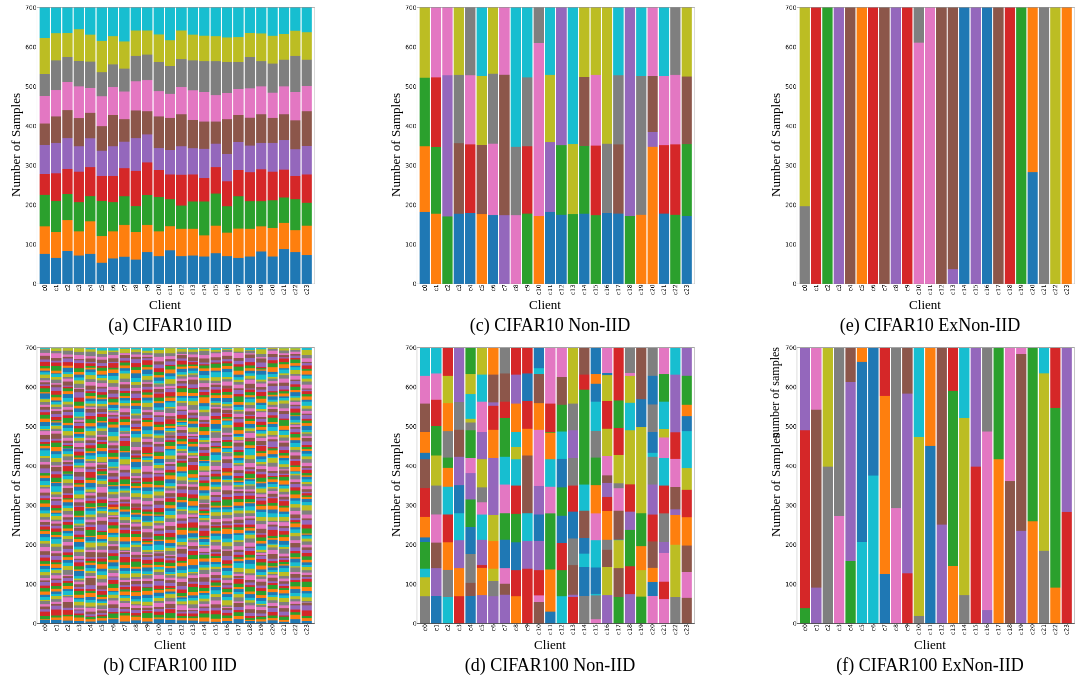}
    \caption{Data distribution of the first 24 clients in the CIFAR10 and CIFAR100 datasets. }
    \label{fig:cifar}
\end{figure*}

The CIFAR10 and CIFAR100 datasets are not separated for FL originally, we need to divide these two datasets synthetically. While the test sets are kept at the server for testing the aggregated model, we divide the training set of each dataset into 64 disjoint partitions with an equal number of samples in three different ways to simulate three scenarios of heterogeneity that are IID, Non-IID, and Extreme Non-IID (ExNon-IID). The IID strategy adapts independent and random division, as shown in Fig. \ref{fig:cifar}(a) and \ref{fig:cifar}(b), the data distribution in each client is basically the same. The Non-IID and ExNon-IID strategies use biased divisions proposed in \cite{fedavg, fl-noniid}. Specifically, the whole dataset is sorted according to the labels and divided into different chunks, then these chunks are randomly assigned to different clients. The number of chunks affects the degree of heterogeneity across clients. As shown in Fig. \ref{fig:cifar}(c)-(f), while each client in Non-IID contains approximately four and ten data classes in CIFAR10 and CIFAR100, respectively, each client in ExNon-IID contains only one and two data classes in CIFAR10 and CIFAR100 respectively, which simulates the extreme data heterogeneity across clients. 

\subsubsection{Model Architecture}
Following previous works \cite{fedavg, leaf}, we study a popular CNN model designed for image classification tasks, called CNN3 on the two datasets. The model only includes two 5x5 convolution layers (the first with 32 channels, the second with 64), each followed by a ReLU activation function and a 2x2 max pooling. After that, one fully connected layer with 512 units and ReLu activation is added, followed by a softmax layer as a classifier. The number of output units is 10 for CIFAR10 and 100 for CIFAR100. By its simple architecture, the model does not need massive resources for training, making it suitable for deployment on IoT-Edge devices. 

\subsection{Hardware and Software Specifications}
\begin{figure}[!ht]
    \centering
    \includegraphics[keepaspectratio, width=0.48\textwidth]{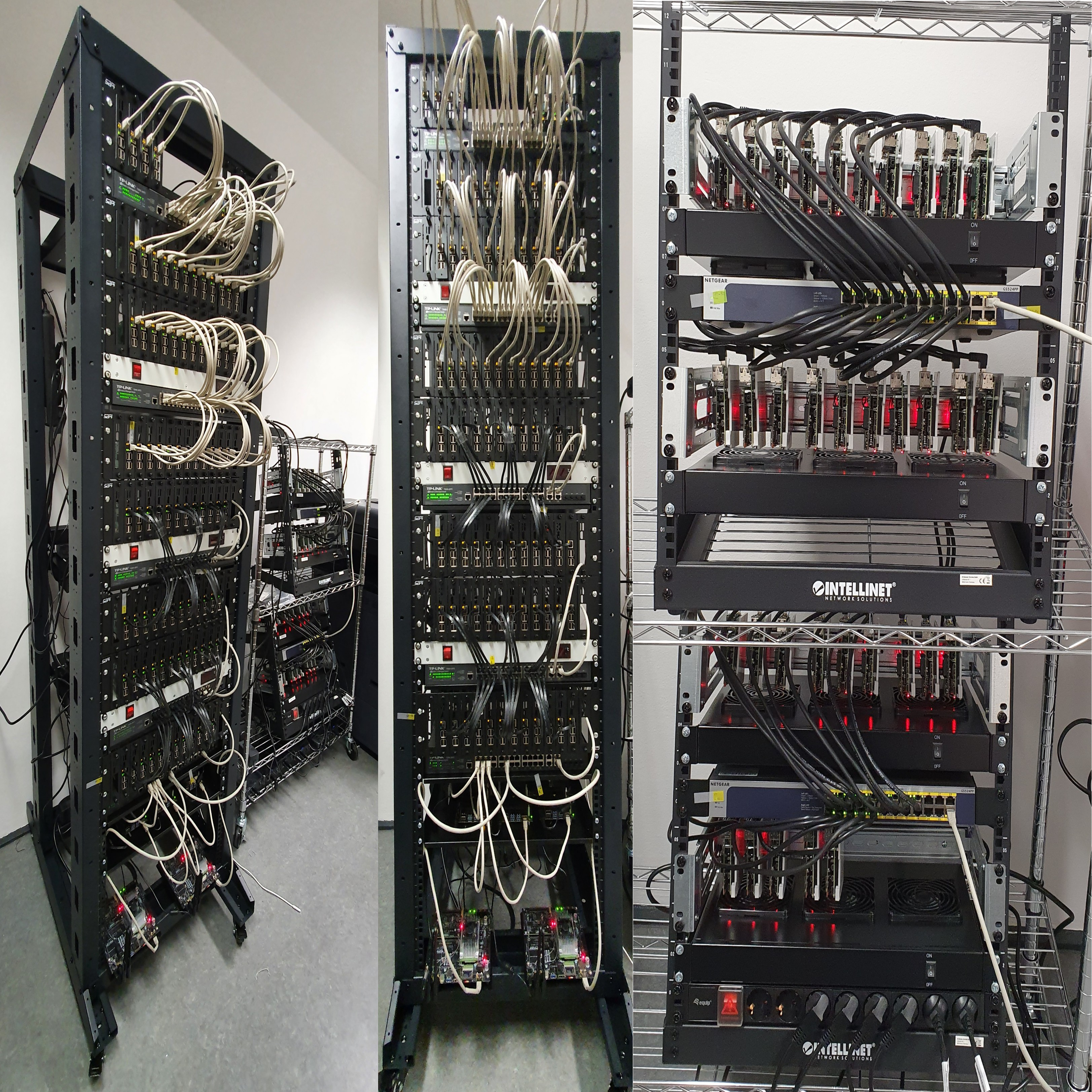}
    \caption{IoT-Edge Federated Learning Testbed. }
    \label{fig:infras}
\end{figure}

In the past few years, many IoT-Edge devices have entered the market with different prices and abilities. In this work, we use the most popular ones such as Pi3, Pi4, Nano, and TX2. Different types of devices with different generations have different resources and processing capabilities. A diverse pool of devices helps us more accurately represent the real world. Our devices are connected to a workstation, which is used as the server, via a network of IoT-Edge devices and switches. Fig.~\ref{fig:infras} is a snapshot of our infrastructure. In more detail, Table~\ref{tab:hardware} provides specifications of these devices, and the server machine and simulation machine are also described.

For software specifications, we use the PyTorch \cite{PyTorch} framework version 1.13.1 to implement deep learning components and use the Flower \cite{Flower} framework version 1.11.0 FedAvg algorithm. Additionally, we use Docker technology to create a separate container on each device to perform local training. 

\begin{table*}[!ht]
\setlength{\tabcolsep}{7.33pt}
\renewcommand{\arraystretch}{1.1}
\caption{Hardware Specifications. }
\label{tab:hardware}
\begin{tabular}[t]{|l|r|r|r|r|}
\hline
Machine     & Memory                & CPU                                                                                       & GPU                & Connectivity \\ \hline
Pi 3        & 1GB LPDDR2 \hspace{0.17em} 900 MHz (32 bit)    & 4-Core ARM-A53 1.20 GHz                                                                   & --                 & 100 Mbps     \\ \hline
Pi 4        & 8GB LPDDR4 3200 MHz (32 bit)  & 4-Core ARM-A72 1.50 GHz                                                                   & --                 & 1 Gbps       \\ \hline
Jetson Nano & 4GB LPDDR4 1600 MHz (64 bit)   & 4-Core ARM-A57 1.43 GHz                                                                   & Maxwell \hspace{0.1em} 4GB x1     & 1 Gbps       \\ \hline
Jetson TX2  & 8GB LPDDR4 1866 MHz (64 bit)  & \begin{tabular}[t]{@{}r@{}}4-Core ARM-A57 2.00 GHz\\ \& 2-Core \hspace{0.6em} Denver2 2.00 GHz\end{tabular} & Pascal \hspace{0.1em} 8GB x1      & 1 Gbps       \\ \hline
Server      & 2048GB DDR4 2666 MHz (64 bit) & Intel Xeon Gold 5117 2.00 GHz                                                             & Tesla V100 16GB x2 & 1 Gbps       \\ \hline
Simulation  & 256GB LPDDR4 2666 MHz (64 bit) & Intel Xeon Gold 6242 2.80 GHz                                                             & RTX 3090 24GB x4   & --           \\ \hline
\end{tabular}
\end{table*}

\subsection{Evaluation Metrics}
In this study, we use a comprehensive set of metrics to characterize and quantify the impact of heterogeneity factors on the behaviors of FL implementation in realistic environments. Specifically, test accuracy and convergence speed are used to evaluate the learning performance. Averaged training time, memory, and GPU/CPU utilization are used to measure computational costs. Finally, we use the averaged model update (local and global) exchange time between the clients and the aggregation server to measure the communication cost. Table \ref{tab:metrics} provides concise definitions of all our used metrics. 

\begin{table*}[!ht]
\setlength{\tabcolsep}{2.36pt}
\renewcommand{\arraystretch}{1.1}
\caption{Evaluation Metric Definitions. }
\label{tab:metrics}
\begin{tabular}{|lll|l|l|}
\hline
\multicolumn{3}{|l|}{Metrics}                                                                                                                                               & Unit       & Definition                                                                                                           \\ \hline
\multicolumn{2}{|l|}{\multirow[t]{2}{*}{Performance}}                                                     & Test Accuracy                                                      & Percentage         & Accuracy of the global model on the test set at the server                                                           \\ \cline{3-5} 
\multicolumn{2}{|l|}{}                                                                                 & \begin{tabular}[t]{@{}l@{}}Convergence\\ Speed\end{tabular}        & No. Rounds & \begin{tabular}[t]{@{}l@{}}The number of communication rounds that\\ the global model needs to converge\end{tabular} \\ \hline
\multicolumn{1}{|l|}{\multirow[t]{4}{*}{Cost}} & \multicolumn{1}{l|}{\multirow[t]{3}{*}{Computational Cost}} & \begin{tabular}[t]{@{}l@{}}Avg Training\\ Time\end{tabular}        & Second     & Average local training time per round for all clients                                                                \\ \cline{3-5} 
\multicolumn{1}{|l|}{}                      & \multicolumn{1}{l|}{}                                    & \begin{tabular}[t]{@{}l@{}}Avg Memory\\ Utilization\end{tabular}   & Percentage & Average memory utilization during training for all clients                                                     \\ \cline{3-5} 
\multicolumn{1}{|l|}{}                      & \multicolumn{1}{l|}{}                                    & \begin{tabular}[t]{@{}l@{}}Avg GPU/CPU\\ Utilization\end{tabular}  & Percentage & Average GPU/CPU utilization during training for all clients                                                    \\ \cline{2-5} 
\multicolumn{1}{|l|}{}                      & \multicolumn{1}{l|}{Communication Cost}                  & \begin{tabular}[t]{@{}l@{}}Avg Update Exchange\\ Time\end{tabular} & Second     & \begin{tabular}[t]{@{}l@{}}Averaged time interval per round when\\ clients send the model to the server until receiving it back\end{tabular} \\ \hline
\end{tabular}
\end{table*}

\subsection{Experiments Setup}

\subsubsection{Behaviors of On-Device FL Implementation (RQ1)}
\label{sec:setup-1}
First of all, we conduct a baseline experiment on the simulation. Particularly, we simulate eight clients in which each client holds one of the first eight partitions (12.5 \% of total partitions) in the CIFAR10 IID dataset. For the training settings, we train a simple CNN3 model described above for 500 communication rounds, at each round, the model is trained for 2 local epochs at the clients, SGD optimizer is used with a learning rate of 0.01, and the batch size is set to 16. To answer the RQ1 described in Section \ref{sec:objectives_RQ}, we then turn the simulation environment in the above experiment into realistic environments by sequentially using eight Pi3s, eight Pi4s, and eight Nanos as clients. These devices are connected to a server machine via ethernet connections. For comparison, all training settings are maintained as in the baseline. We use all metrics defined in Table \ref{tab:metrics} to describe the behaviors of FL implementation. The results and conclusions are shown in Section \ref{sec:rq1-res}. 

\subsubsection{Impact of Single Factor (RQ2)}
\begin{figure*}[!ht]
    \centering
    \includegraphics[keepaspectratio, width=0.98\textwidth]{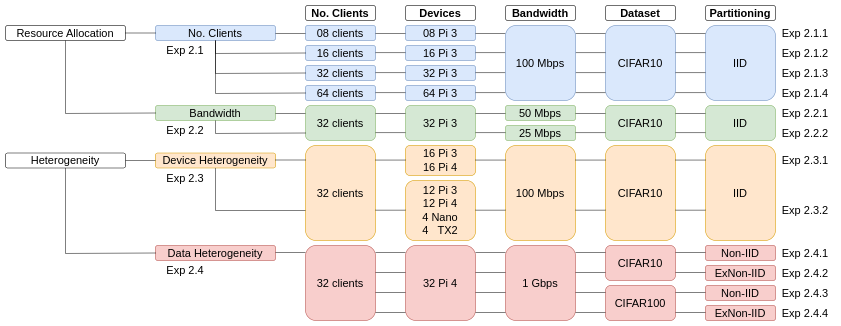}
    \caption{Experiments Setup for Studying the Impact of Single Factor (RQ2). }
    \label{fig:Exps-rq2}
\end{figure*}

For the RQ2, we consider two critical factors in FL, namely resource allocation and heterogeneity. Resource allocation includes the number of participating clients and the connection’s communication bandwidth, and heterogeneity includes device heterogeneity and data heterogeneity (statistical heterogeneity). To explore the impact of these factors, we conduct extensive experiments that are shown in detail in Fig. \ref{fig:Exps-rq2}. Training settings are the same as in the baseline experiment in RQ1. By conducting experiments defined in Fig. \ref{fig:Exps-rq2}, we can observe what happens when the number of participating clients increases, the communication bandwidth is saturated, and when intrinsic heterogeneity is introduced across clients. The results and conclusions for RQ2 experiments are provided in Section \ref{sec:rq2-res}. 

\subsubsection{Impact of Combined Factors (RQ3)}
\begin{figure*}[!ht]
    \centering
    \includegraphics[keepaspectratio, width=0.98\textwidth]{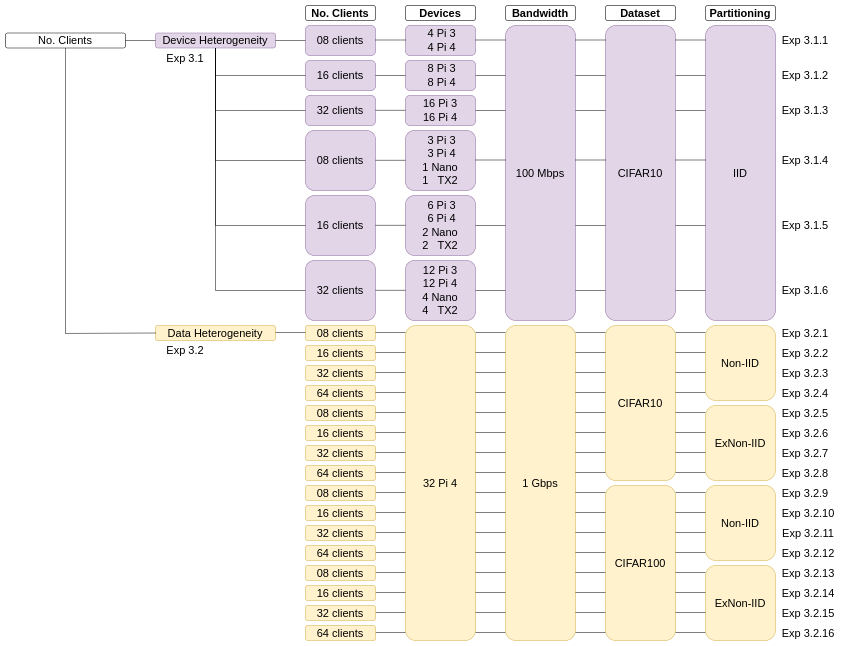}
    \caption{Experiments Setup for Studying the Impact of Combined Factors (RQ3). }
    \label{fig:Exps-rq3}
\end{figure*}

After observing the impact of resource allocation and heterogeneity individually by addressing RQ2, we aim to explore more realistic scenarios where these two factors appear simultaneously. First, we vary the number of participating clients and increase the degree of heterogeneity in client devices concurrently. Second, we still vary the number of participating clients in different data heterogeneity settings (IID, Non-IID, and ExNon-IID) to observe the accuracy and convergence speed. Fig. \ref{fig:Exps-rq3} shows these experiments in detail. Additionally, training settings are the same as in the baseline experiment in RQ1. By conducting these experiments, we expect to gain more valuable insights beyond those gained from the RQ2. Also, we aim to figure out the dominant factors towards the behaviors of FL in real-device deployment. The results and conclusions for RQ3 experiments are provided in Section \ref{sec:rq3-res}. 

\section{Experimental Results}
\label{sec:results}

\subsection{Behaviors of On-Device FL Implementation (RQ1)}
\label{sec:rq1-res}
Table \ref{tab:rq1} provides detailed results of experiments in RQ1 where we compare real-device FL implementations to the baseline of simulation. Details of the experimental setup are described in \ref{sec:setup-1}. All four experiments use the same eight partitions of the CIFAR10 IID dataset and the same training details, it is reasonable that test accuracy and convergence speed in these experiments are consistent. In terms of computational cost, training time exponentially increases when we change the devices from TX2 and Nano to Pi4, then Pi3. From resource utilization, Pi3 devices seem to be overloaded when training a small model like CNN3, while Nano devices can handle the task easier due to the support of GPU. Additionally, update exchange time roughly doubles when we change the devices from Nano to Pi4, then Pi3. These observations raise a need for more efficient FL frameworks which are suitable for low-end devices like Pi3, and even for weaker, lower-cost IoT devices or sensors which were introduced more and more with extremely limited computational capacity. 


\begin{table*}[!ht]
\setlength{\tabcolsep}{4.39pt}
\renewcommand{\arraystretch}{1.1}
\caption{Behaviors of On-Device FL Implementation (RQ1). }
\label{tab:rq1}
\begin{tabular}{|l|rr|rrrr|}
\hline
\multirow[t]{3}{*}{Hardware} & \multicolumn{2}{l|}{\multirow[t]{2}{*}{Performance}}                                                                                  & \multicolumn{4}{l|}{Operation Cost}                                                                                                                                                                                                                                                                                                                                                     \\ \cline{4-7} 
                             & \multicolumn{2}{l|}{}                                                                                                              & \multicolumn{3}{l|}{Computational Cost}                                                                                                                                                                                                                                         & \multicolumn{1}{l|}{Communication Cost}                                                     \\ \cline{2-7} 
                             & \multicolumn{1}{l|}{Test Accuracy} & \multicolumn{1}{l|}{\begin{tabular}[t]{@{}l@{}}Convergence\\ Speed (no. rounds)\end{tabular}} & \multicolumn{1}{l|}{\begin{tabular}[t]{@{}l@{}}Avg Training\\ Time (s)\end{tabular}} & \multicolumn{1}{l|}{\begin{tabular}[t]{@{}l@{}}Avg Memory\\ Utilization (\%)\end{tabular}} & \multicolumn{1}{l|}{\begin{tabular}[t]{@{}l@{}}Avg GPU/CPU\\ Utilization (\%)\end{tabular}} & \multicolumn{1}{l|}{\begin{tabular}[t]{@{}l@{}}Avg Update Exchange\\ Time (s)\end{tabular}} \\ \hline
Pi 3                       & \multicolumn{1}{r|}{0.662}         & 322                                                                                           & \multicolumn{1}{r|}{161.148}                                                         & \multicolumn{1}{r|}{40.851}                                                                & \multicolumn{1}{r|}{-- / 73.188}                                                            & 52.471                                                                                      \\ \hline
Pi 4                       & \multicolumn{1}{r|}{0.664}         & 338                                                                                           & \multicolumn{1}{r|}{22.739}                                                          & \multicolumn{1}{r|}{11.414}                                                                & \multicolumn{1}{r|}{-- / 42.312}                                                            & 25.523                                                                                      \\ \hline
Nano                       & \multicolumn{1}{r|}{0.660}         & 339                                                                                           & \multicolumn{1}{r|}{5.211}                                                           & \multicolumn{1}{r|}{77.213}                                                                & \multicolumn{1}{r|}{56.177 / 11.309}                                                        & 12.915                                                                                      \\ \hline
Simulation                   & \multicolumn{1}{r|}{0.667}         & 314                                                                                           & \multicolumn{1}{r|}{1.524}                                                           & \multicolumn{1}{r|}{13.363}                                                                & \multicolumn{1}{r|}{11.118 / \hspace{0.1em} 0.578}                                                         & 3.949                                                                                       \\ \hline
\end{tabular}
\end{table*}

\subsection{Impact of Single Factor On FL Implementation (RQ2)}
\label{sec:rq2-res}

In this set of experiments, we observe the results of experiments in RQ2 and analyze what happens when the number of participating clients increases, the communication bandwidth is constrained, and when intrinsic heterogeneity is introduced across clients. 

\subsubsection{Impact of the Resource Allocation}
\newblock

\begin{figure}[!ht]
    \raggedleft
    \includegraphics[keepaspectratio, width=0.49\textwidth]{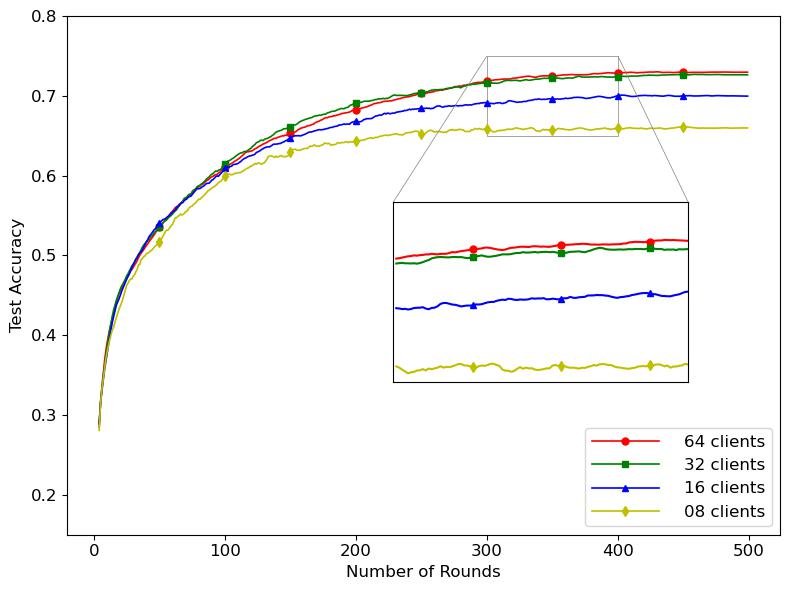}
    \caption{Impact of the Number of Clients on Test Accuracy. }
    \label{fig:num_clients-accuracy}
\end{figure}

\begin{figure}[!ht]
    \raggedleft
    \includegraphics[keepaspectratio, width=0.49\textwidth]{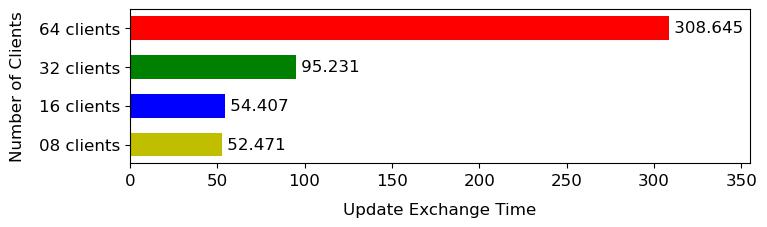}
    \caption{Impact of the Number of Clients on Update Exchange Time. }
    \label{fig:num_clients-comm_time}
\end{figure}

\textbf{Impact of the Number of Clients}. Fig. \ref{fig:num_clients-accuracy} and Fig. \ref{fig:num_clients-comm_time} show the effect of the number of participating clients on the learning performance of communication cost. Generally, increasing the number of clients means more data involved in training the global model, resulting in an improvement in test accuracy. However, this also leads to a high diversity across client model parameters which can slow down the convergence process. We also observe that \emph{when the number of clients increases from 32 to 64, the improvement in test accuracy is \textbf{negligible}, however, the update exchange time goes up dramatically}. From this observation, we can empirically verify an assumption that more participating clients do not guarantee better accuracy but can lead to large congestion in communication and increase the update exchange time. In this setting, it is easy to observe that 32 is the optimal number of participating clients. Therefore, we only use 32 clients in the remaining experiments in RQ2. 


\begin{figure}[!ht]
    \raggedleft
    \includegraphics[keepaspectratio, width=0.49\textwidth]{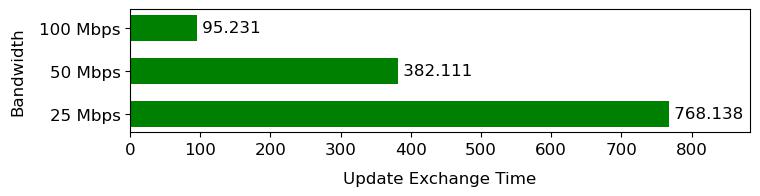}
    \caption{Impact of the Bandwidth on Update Exchange Time. }
    \label{fig:bandwidth-comm_time}
\end{figure}

\textbf{Impact of the Communication Bandwidth}. Next, we investigate the effect of connection bandwidth on update exchange time. One interesting point obtained from Fig. \ref{fig:bandwidth-comm_time} is that update exchange time increase linearly when we decrease the bandwidth. Specifically, when we halve the bandwidth from 100Mbps to 50Mbps, the update exchange time increases approximately by 4 times. Furthermore, it increases about by 8 times when the bandwidth is constrained four times from 100Mbps to 25Mbps. This observation promotes FL algorithms that are suitable for low-bandwidth systems. 

\subsubsection{Impact of the Heterogeneity}
\newblock

\textbf{Impact of the Device Heterogeneity}. Following the experiments in Fig. \ref{fig:Exps-rq2}, we investigate the impact of heterogeneity across client devices. From Table \ref{tab:device-heterogeneity} below, we can observe that in a federation of heterogeneous devices, more powerful devices such as Nano or TX2 only need a couple of seconds to finish local training while weaker devices like Pi3 and Pi4 need much longer. However, in a naive FedAvg framework, the server needs to wait for all clients regardless of their strengths which is the reason why the update exchange time of more powerful devices is higher than weaker devices, this diminishes all benefits that high-end devices bring. This observation suggests \emph{a need for better client selection strategies based on the client's computational power in realistic systems to leverage the presence of high-end devices}. 

\begin{table}[!ht]
\setlength{\tabcolsep}{4.13pt}
\renewcommand{\arraystretch}{1.1}
\caption{Impact of the Device Heterogeneity. }
\label{tab:device-heterogeneity}
\begin{tabular}{|l|r|r|r|}
\hline
Exps                       & \multicolumn{1}{l|}{Devices} & \multicolumn{1}{l|}{\begin{tabular}[t]{@{}l@{}}Avg Training\\ Time (s)\end{tabular}} & \multicolumn{1}{l|}{\begin{tabular}[t]{@{}l@{}}Avg Update Exchange\\ Time (s)\end{tabular}} \\ \hline
Exp 2.1.3                  & 32 Pi 3                      & 161.872                                                                              & 87.090                                                                                      \\ \hline
\multirow[t]{2}{*}{Exp 2.3.1} & 16 Pi 3                      & 166.641                                                                              & 72.826                                                                                      \\ \cline{2-4} 
                           & 16 Pi 4                      & 22.715                                                                               & 216.448                                                                                     \\ \hline
\multirow[t]{4}{*}{Exp 2.3.2} & 12 Pi 3                      & 170.227                                                                              & 65.391                                                                                      \\ \cline{2-4} 
                           & 12 Pi 4                      & 22.687                                                                               & 215.620                                                                                     \\ \cline{2-4} 
                           & 4 Nano                       & 4.971                                                                                & 233.224                                                                                     \\ \cline{2-4} 
                           & 4 \hspace{0.001em} TX2                        & 4.126                                                                                & 234.161                                                                                     \\ \hline
\end{tabular}
\end{table}

\begin{figure*}[!ht]
    \centering
    \includegraphics[keepaspectratio, width=0.95\textwidth]{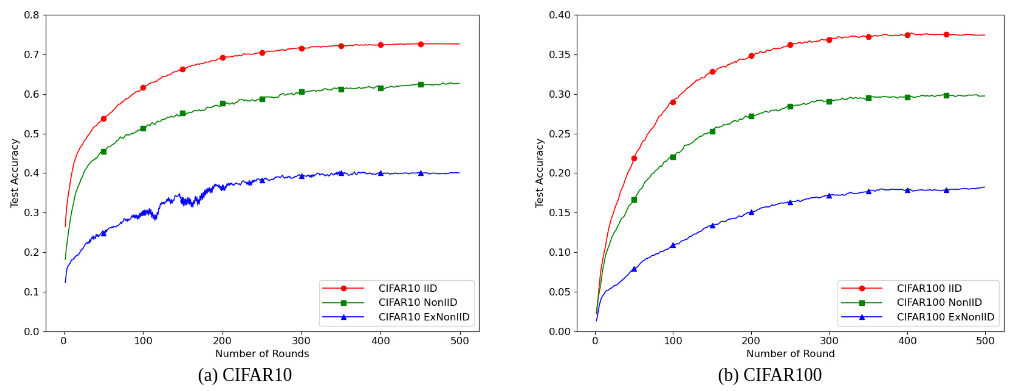}
    \caption{Impact of the Data Heterogeneity. }
    \label{tab:data-heterogeneity}
\end{figure*}

\textbf{Impact of the Data Heterogeneity}. Heterogeneous data or distribution shift is the most challenging issue in FL. Most existing works on this issue only consider conventional Non-IID data scenarios. As discussed above, in this study, we further explore extreme cases of heterogeneity, i.e., ExNon-IID. Figs. \ref{tab:data-heterogeneity}(a) and \ref{tab:data-heterogeneity}(b) to show the effect of data heterogeneity on FL for CIFAR10 and CIFAR100 datasets, respectively. As observed from these results, ExNon-IID scenarios degrade the accuracy on test sets significantly compared to IID and Non-IID cases. Additionally, ExNon-IID scenarios tend to cause some fluctuation periods during training and slow down the convergence process. This suggests that the development of FL algorithms needs to tackle not only Non-IID cases but also ExNon-IID. 

In summary, we have figured out that increasing the number of participating clients generally leads to an improvement in accuracy due to the increase in data samples used for training. However, when we substantially increase the number of clients (i.e., from 32 to 64), the improvement is not significant but the update exchange time goes up dramatically. Moreover, the data heterogeneity also affects the global model’s accuracy significantly, especially in ExNon-IID cases. Besides heterogeneity in labels of local datasets, other types of data heterogeneity such as quantity heterogeneity or distribution heterogeneity are also important and might degrade the model’s accuracy much further, however, these types of data heterogeneity are still under-explored. In addition, the update exchange time is linearly affected by communication bandwidth. Also, we show that better client selection strategies are essential when dealing with heterogeneous devices to leverage the presence of high-end devices and reduce the update exchange time. However, \emph{it is quite challenging on a real deployment when the distributions of computing power and data are not known as a prior and can not be simulated in a controlled setting}.

\subsection{Impact of Combined Factors On FL Implementation (RQ3)}
\label{sec:rq3-res}


This part reports the experimental results of RQ3 and draws insights when two factors, resource allocation, and heterogeneity, appear simultaneously. Also, we aim to figure out dominant factors towards the FL behaviors in real-device deployment. 

\begin{figure}[!ht]
    \raggedleft
    \includegraphics[keepaspectratio, width=0.49\textwidth]{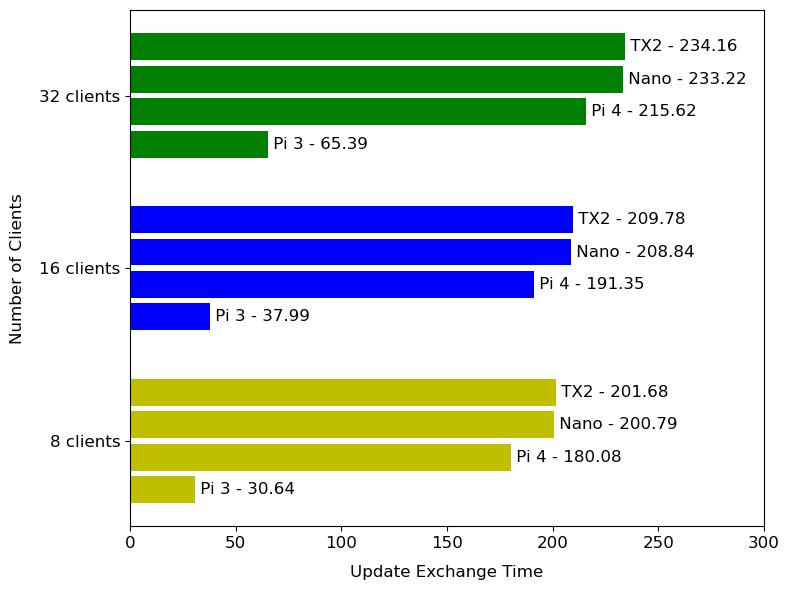}
    \caption{Combined Impact of the Number of Clients and Device Heterogeneity on Update Exchange Time. }
    \label{fig:combined-comm-time}
\end{figure}

\textbf{Combined Impact of the Number of Clients and Device Heterogeneity}. We focus on investigating the effect of the number of clients and device heterogeneity across clients on the update exchange time. Fig. \ref{fig:combined-comm-time} shows the average update exchange time of each type of device used in experiments 3.1.4 to 3.1.6. By comparing these results with results in Fig. \ref{fig:num_clients-comm_time} and Table \ref{tab:device-heterogeneity}, we can draw a fascinating insight that with the same number of clients, heterogeneity in the federation can help reduce the overall update exchange time, and this gap seems more significant with a smaller number of clients. Unlike in homogenous scenarios where clients mostly finish local training and update their local models to the server simultaneously, which causes considerable congestion, in heterogeneous scenarios, clients with more powerful devices complete their work earlier, followed by weaker devices sequentially. This helps reduce the congestion in communication. These observations also suggest that a large number of clients and the congestion have a significantly negative effect on the update exchange time and raise a need for novel FL algorithms capable of handling situations with massive clients. 

\begin{figure*}[!ht]
    \centering
    \includegraphics[keepaspectratio, width=0.965\textwidth]{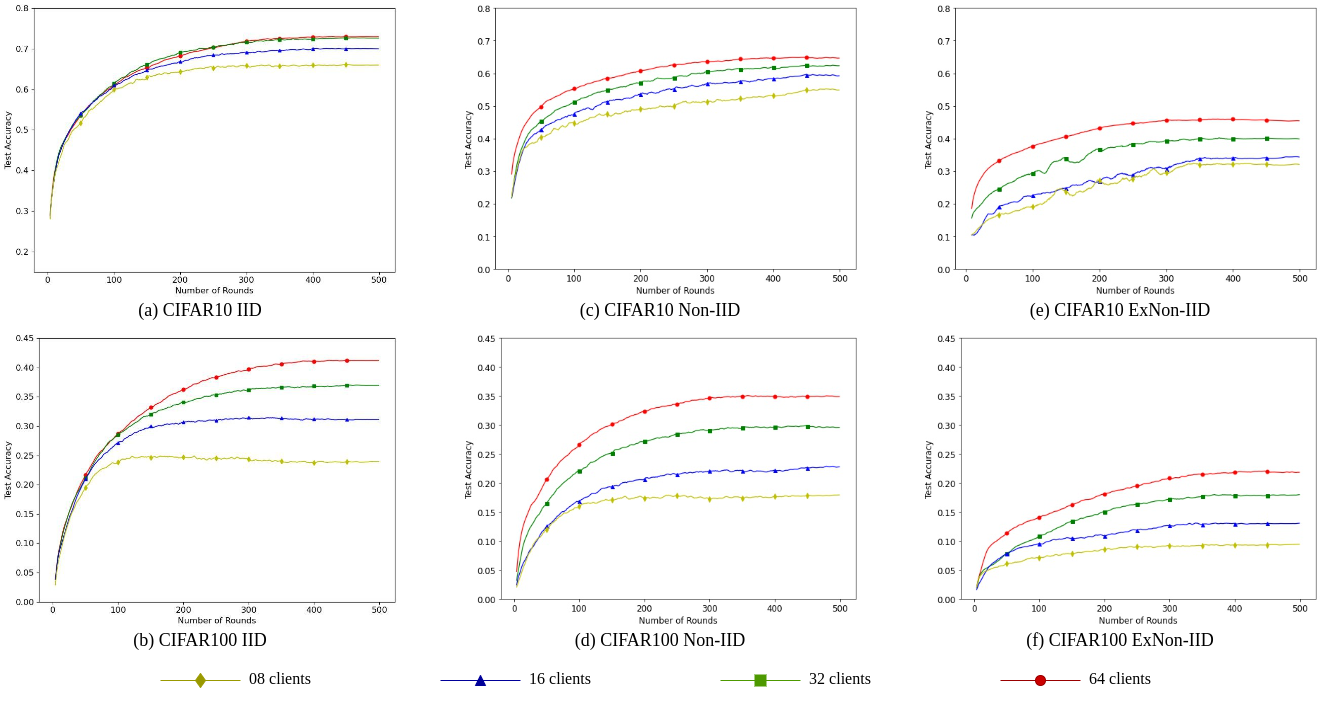}
    \caption{Combined Impact of the Number of Clients and Data Heterogeneity. }
    \label{fig:combined-acc}
\end{figure*}

\textbf{Combined Impact of the Number of Clients and Data Heterogeneity}. We continue to study simultaneously the effect of the number of clients and data heterogeneity. Fig. \ref{fig:combined-acc} shows the test accuracy of the global model in experiments 3.2.1 to 3.2.16. From Fig. \ref{fig:combined-acc}(a), \ref{fig:combined-acc}(c), and \ref{fig:combined-acc}(e), we can see that when increasing the number of clients from 32 to 64, the improvement in IID case is negligible. However, the improvement is more significant in cases of Non-IID and ExNon-IID which means that a large number of participating clients is essential in heterogenous data scenarios. Moreover, the negative effect of ExNon-IID data on the more challenging dataset, CIFAR100, seems more serious. Therefore, we can conclude that \emph{ data heterogeneity is the most dominant factor in the model’s test accuracy, especially in challenging datasets}. 

In summary, we have figured out that the communication congestion caused by a large number of clients has a significant negative effect on the update exchange time. However, increasing the number of clients leads to improvements in accuracy, especially in heterogenous data scenarios. Also, data heterogeneity is the most dominant factor that affects the model’s test accuracy, especially in challenging datasets. Going beyond the fundamental image classification task, data heterogeneity might further hurt the model’s performance in other advanced tasks, such as object detection or segmentation, which are under-explored in current literature. Interestingly, we also observe that some homogeneous devices can behave differently. This may be caused by various implicit factors such as power supply, network conditions, hardware and software variations, or user behavior.

\section{Discussions}
\label{sec:discussions}
In this section, we first discuss the practicality of FL on IoT-Edge devices (based on our experimental results) and then discuss other essential factors to consider while designing an FL system for IoT devices. 

\subsection{Practicality of FL on IoT-Edge Devices}
FL requires local processing on the device, which can be challenging on lightweight devices with limited processing power. In addition, storing the model updates locally can be challenging due to the limited storage capacity. Another challenge is the unreliable connectivity of IoT devices. Federated learning requires a stable and reliable network connection for devices to communicate with each other and the aggregation server. However, IoT-Edge devices are often deployed in remote locations with limited network connectivity. 

In this study, we observed that the practicality of FL on IoT-Edge devices depends on combined effects from various factors such as device availability (number of participating clients), communication constraints (bandwidth availability), and heterogeneity of data (data distribution) and devices (computational capability and hardware configuration). These factors are interdependent and affect each other, and hence, a comprehensive analysis of the practicality of FL on IoT devices should consider all these factors together. For example, the computational capability of devices can affect communication overhead, as devices with lower computational capability may take longer to process and transmit data, resulting in higher communication latency and overhead. Similarly, the heterogeneity of devices can affect the robustness of FL algorithms, as the presence of devices with varying characteristics can introduce heterogeneity in the data and make it challenging to train accurate models.

To address the processing power and storage capacity issues, we need to design models that are optimized for lightweight devices and implement compression or distillation techniques to reduce the size of the updates. There is also a need to implement techniques such as asynchronous updates and checkpointing to ensure that the training process can continue even when devices are disconnected due to network connectivity issues. 

\subsection{Other Considerable Factors}
Besides the factors studied in this work, it is essential to consider other factors that can cause IoT devices not to perform well in FL, such as the power supply of devices and specifications of memory cards, and the performance of the aggregation server when designing FL systems. 

\subsubsection{Power Supply} The amount of power available to the device can impact its processing capability. If the device has a limited power supply, it may not be able to perform complex computations or transmit large amounts of data efficiently. Furthermore, the quality and reliability of the power supply can affect the device's stability and longevity. Power surges or outages can cause damage to the device's components, leading to reduced performance and potentially even complete failure. As shown in \cite{fl-design}, when the battery life of the devices decreased, the accuracy of the global model also decreased significantly. Hence, it is crucial to ensure that devices used in FL have access to a reliable power supply with sufficient capacity to handle the demands of the learning process. 

\subsubsection{Memory Card Usage} The speed and capacity of the memory card can indirectly affect the overall performance of the IoT device itself. If the memory card is slow or has limited capacity, it may result in slower data processing and storage, slowing down the overall FL process. Also, the reliability and durability of the memory card can impact FL performance. For instance, if the memory card fails or becomes corrupted, it can result in the loss of data, which can negatively impact the accuracy and effectiveness of the FL model. 

\subsubsection{Performance of the Aggregation Server} The performance of the aggregation server is crucial to the success of the FL process and can bring a significant impact on the participating IoT devices. The aggregation server needs to have sufficient computational resources to process the incoming model updates from IoT devices. If the server is overloaded, this can cause delays or even crashes in the system, affecting the IoT devices involved. This can be particularly problematic if the IoT devices have limited resources themselves, as they may not be able to handle the increased workload. 

\section{Conclusions and Future Works}
\label{sec:conclusion_limitation_futurework}
The results of our experiment have revealed several important findings: (1) our simulation of FL has shown that it can be a valuable tool for algorithm testing and evaluation, but its effectiveness in accurately representing the reality of IoT-Edge deployment is very limited, (2) the disparity in computational resources among IoT devices can significantly impact the update exchange time, and (3) data heterogeneity is the most dominant factor in the presence of other factors, especially working in tandem with computation and network factors. 

Moving forward, several areas could be explored to expand on the findings of this study. Firstly, considering the diversity of devices used in FL, it would be valuable to test the approach on a more comprehensive range of devices with different hardware, operating systems, and network connections to ensure the effectiveness and robustness of the approach. Secondly, the dataset selection process used for training the FL model could be further optimized to increase accuracy and efficiency and ensure that the results represent all potential use cases. Additionally, to expand the scope of the study's findings, exploring other FL algorithms beyond the standard FedAvg algorithm could be beneficial. These alternative algorithms could be better suited for specific scenarios or applications and may provide insights into how to improve the performance of FL in IoT-Edge devices. Lastly, the study may miss out on the potential benefits of other FL algorithms that are better suited for specific scenarios or applications. For instance, FedProx \cite{fedprox} is designed to handle heterogeneity in data across devices and can improve the convergence rate of the FL process. It is important to note that these future improvements do not affect the objectives and scopes of the current study. 

Particularly, we plan to extend our study to a broader range of scenarios by examining the impact of varying network conditions, communication protocols, and resource usage of FL. In addition, we want to conduct a comprehensive analysis to measure the resource consumption of FL, including battery life and network bandwidth usage. We also want to focus on real-world applications of FL on IoT devices, including developing FL-based solutions for specific IoT use cases such as environmental monitoring, predictive maintenance, and evaluating their performance in realistic environments. 

\bibliographystyle{unsrt}
\bibliography{refs}

\begin{thebibliography}{10}

\bibitem{iotdevices}
{Internet of Things (IoT) And Non-IoT Active Device Connections Worldwide From
  2010 to 2025}.
\newblock
  https://www.statista.com/statistics/1101442/iot-number-of-connected-devices-worldwide.

\bibitem{100barms}
{Enabling Mass IoT Connectivity as Arm Partners Ship 100 Billion Chips}.
\newblock
  https://community.arm.com/arm-community-blogs/b/internet-of-things-blog/posts/enabling-mass-iot-connectivity-as-arm-partners-ship-100-billion-chips.

\bibitem{DBLP:journals/cacm/Groger21}
Christoph Gr{\"{o}}ger.
\newblock {There Is No AI Without Data}.
\newblock {\em Commun. {ACM}}, 64(11):98--108, 2021.

\bibitem{GDPR}
Axel von dem~Bussche Paul~Voigt.
\newblock {\em {The EU General Data Protection Regulation (GDPR)}}.
\newblock Springer Cham, 2017.

\bibitem{fedavg}
H.~Brendan McMahan, Eider Moore, Daniel Ramage, Seth Hampson, and
  Blaise~Agüera y~Arcas.
\newblock {Communication-Efficient Learning of Deep Networks From Decentralized
  Data}.
\newblock In {\em {Proceedings of the 20th International Conference on
  Artificial Intelligence and Statistics, Aistats 2017}}, 2017.

\bibitem{Survey-Concept}
Qiang Yang, Yang Liu, Tianjian Chen, and Yongxin Tong.
\newblock {Federated Machine Learning: Concept and Applications}.
\newblock {\em {ACM Trans. on Intelligent Systems and Technology}}, 2019.

\bibitem{Survey-Vision}
Qinbin Li, Zeyi Wen, Zhaomin Wu, Sixu Hu, Naibo Wang, Yuan Li, Xu~Liu, and
  Bingsheng He.
\newblock {A Survey on Federated Learning Systems: Vision, Hype and Reality for
  Data Privacy and Protection}.
\newblock {\em {IEEE Trans. on Knowledge and Data Engineering}}, 2021.

\bibitem{fedadapt}
Di~Wu, Rehmat Ullah, Paul Harvey, Peter Kilpatrick, Ivor Spence, and Blesson
  Varghese.
\newblock {FedAdapt: Adaptive Offloading for IoT Devices in Federated
  Learning}.
\newblock {\em {IEEE Internet of Things Journal}}, 9(21):20889--20901, 2022.

\bibitem{fedmsa}
Rui Sun, Yinhao Li, Tejal Shah, Ringo W.~H. Sham, Tomasz Szydlo, Bin Qian,
  Dhaval Thakker, and Rajiv Ranjan.
\newblock {FedMSA: Fedmsa: A Model Selection and Adaptation System for
  Federated Learning}.
\newblock {\em {Sensors}}, 22(19), 2022.

\bibitem{ce-fedavg}
Jed Mills, Jia Hu, and Geyong Min.
\newblock {Communication-Efficient Federated Learning for Wireless Edge
  Intelligence in IoT}.
\newblock {\em {IEEE Internet of Things Journal}}, 7(7):5986--5994, 2020.

\bibitem{fedprox}
Tian Li, Anit~Kumar Sahu, Manzil Zaheer, Maziar Sanjabi, Ameet Talwalkar, and
  Virginia Smith.
\newblock {Federated Optimization in Heterogeneous Networks}.
\newblock In I.~Dhillon, D.~Papailiopoulos, and V.~Sze, editors, {\em
  {Proceedings of Machine Learning and Systems}}, pages 429--450, 2020.

\bibitem{fedma}
Hongyi Wang, Mikhail Yurochkin, Yuekai Sun, Dimitris Papailiopoulos, and
  Yasaman Khazaeni.
\newblock {Federated Learning with Matched Averaging}.
\newblock In {\em {International Conference on Learning Representations}},
  2020.

\bibitem{Chen-survey}
Huiming Chen, Huandong Wang, Qingyue Long, Depeng Jin, and Yong Li.
\newblock {Advancements in Federated Learning: Models, Methods, and Privacy}.
\newblock {\em {arXiv}}, 2023.

\bibitem{Liu-survey}
Bingyan Liu, Nuoyan Lv, Yuanchun Guo, and Yawen Li.
\newblock {Recent Advances on Federated Learning: A Systematic Survey}.
\newblock {\em {arXiv}}, 2023.

\bibitem{fl-iot-1}
Dinh~C. Nguyen, Ming Ding, Pubudu~N. Pathirana, Aruna Seneviratne, Jun Li, and
  H.~Vincent Poor.
\newblock {Federated Learning for Internet of Things: A Comprehensive Survey}.
\newblock {\em {IEEE Communications Surveys and Tutorials}}, 23, 2021.

\bibitem{fl-iot-2}
Ahmed Imteaj, Urmish Thakker, Shiqiang Wang, Jian Li, and M.~Hadi Amini.
\newblock {A Survey on Federated Learning for Resource-Constrained IoT
  Devices}.
\newblock {\em {IEEE Internet of Things Journal}}, 9, 2022.

\bibitem{noniid-survey}
Hangyu Zhu, Jinjin Xu, Shiqing Liu, and Yaochu Jin.
\newblock {Federated Learning on Non-iid Data: A Survey}.
\newblock {\em {Neurocomputing}}, 465:371--390, 2021.

\bibitem{noniid-study}
Qinbin Li, Yiqun Diao, Quan Chen, and Bingsheng He.
\newblock {Federated Learning on Non-IID Data Silos: An Experimental Study}.
\newblock In {\em {2022 IEEE 38th International Conference on Data Engineering
  (ICDE)}}, 2022.

\bibitem{personalized-study}
Koji Matsuda, Yuya Sasaki, Chuan Xiao, and Makoto Onizuka.
\newblock {An Empirical Study of Personalized Federated Learning}.
\newblock {\em {arXiv}}, 2022.

\bibitem{leaf}
Sebastian Caldas, Sai Meher~Karthik Duddu, Peter Wu, Tian Li, Jakub Konečný,
  H.~Brendan McMahan, Virginia Smith, and Ameet Talwalkar.
\newblock {LEAF: A Benchmark for Federated Settings}.
\newblock In {\em Workshop on Federated Learning for Data Privacy and
  Confidentiality}, 2019.

\bibitem{fliot}
Tuo Zhang, Chaoyang He, Tianhao Ma, Lei Gao, Mark Ma, and Salman Avestimehr.
\newblock {Federated Learning for Internet of Things}.
\newblock In {\em {Proceedings of the 19th ACM Conference on Embedded Networked
  Sensor Systems}}, SenSys '21, page 413–419, New York, NY, USA, 2021.
  {Association for Computing Machinery}.

\bibitem{cifar}
Alex Krizhevsky.
\newblock {Learning Multiple Layers of Features from Tiny Images}.
\newblock {\em {Science Department, University of Toronto, Tech}}, 2009.

\bibitem{fl-noniid}
Yue Zhao, Meng Li, Liangzhen Lai, Naveen Suda, Damon Civin, and Vikas Chandra.
\newblock {Federated Learning with Non-IID Data}.
\newblock {\em {arXiv}}, 2018.

\bibitem{PyTorch}
Adam Paszke, Sam Gross, Francisco Massa, Adam Lerer, James Bradbury, Gregory
  Chanan, Trevor Killeen, Zeming Lin, Natalia Gimelshein, Luca Antiga, et~al.
\newblock {PyTorch: An Imperative Style, High-Performance Deep Learning
  Library}.
\newblock {\em {Advances in Neural Information Processing Systems}}, 32, 2019.

\bibitem{Flower}
Daniel~J Beutel, Taner Topal, Akhil Mathur, Xinchi Qiu, Javier
  Fernandez-Marques, Yan Gao, Lorenzo Sani, Kwing~Hei Li, Titouan Parcollet,
  Pedro Porto~Buarque de~Gusm{\~a}o, et~al.
\newblock {Flower: A Friendly Federated Learning Research Framework}.
\newblock {\em {arXiv:2007.14390}}, 2020.

\bibitem{fl-design}
Keith Bonawitz, Hubert Eichner, Wolfgang Grieskamp, Dzmitry Huba, Alex
  Ingerman, Vladimir Ivanov, Chloe Kiddon, Jakub Kone{\v{c}}n{\`y}, Stefano
  Mazzocchi, Brendan McMahan, et~al.
\newblock {Towards Federated Learning at Scale: System Design}.
\newblock {\em {Proceedings of Machine Learning and Systems}}, 1:374--388,
  2019.

\end{thebibliography}
\section*{Biography Section}

\begin{IEEEbiography}[{\includegraphics[width=1in,height=1.25in,clip,keepaspectratio]{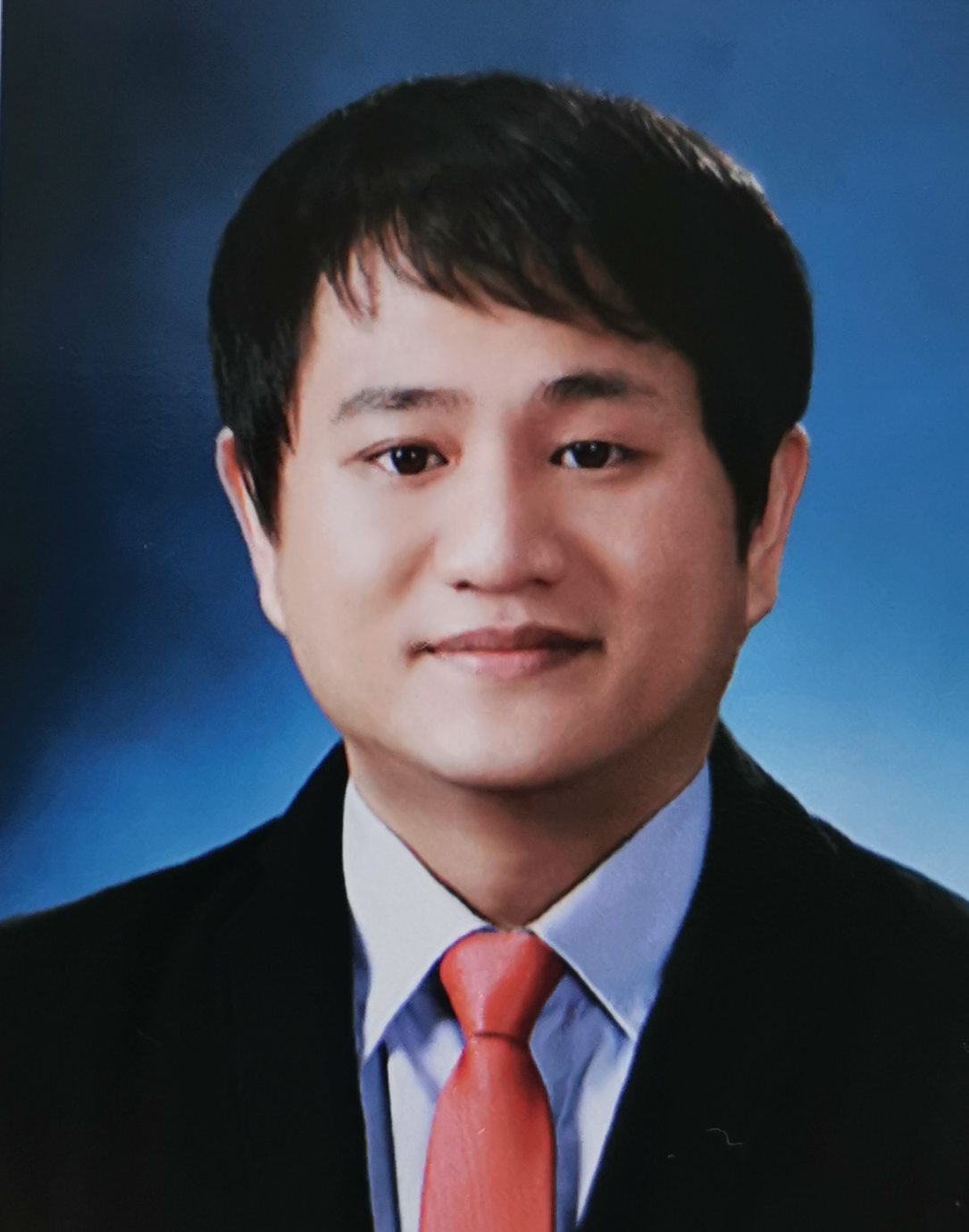}}]{Kok-Seng Wong}(Member, IEEE)
received his first degree in Computer Science (Software Engineering) from the University of Malaya, Malaysia in 2002, and an M.Sc. (Information Technology) degree from Malaysia University of Science and Technology (in collaboration with MIT) in 2004. He obtained his Ph.D. from Soongsil University, South Korea, in 2012. He is currently an Associate Professor in the College of Engineering and Computer Science, VinUniversity. 
To this end, he conducts research that spans areas of security, data privacy, and AI security while maintaining a strong relevance to the privacy-preserving framework. 
\end{IEEEbiography}

\begin{IEEEbiography}[{\includegraphics[width=1in,height=1.25in,clip,keepaspectratio]{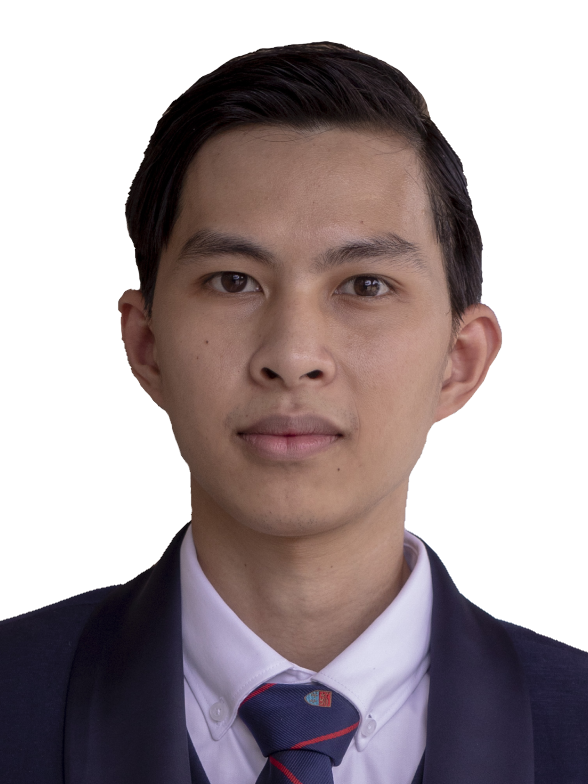}}]{Duc-Manh Nguyen} got a Master in Information Science and Technology from University of Information Science and Technology, North Macedonia. Currently, he is a PhD candidate and a research assistant at the Technical University of Berlin. His research focuses on Robotics and Edge Computing with Machine Learning, partially Cooperative Perception for Autonomous Vehicles.

\end{IEEEbiography}

\begin{IEEEbiography}[{\includegraphics[width=1in,height=1.25in,clip,keepaspectratio]{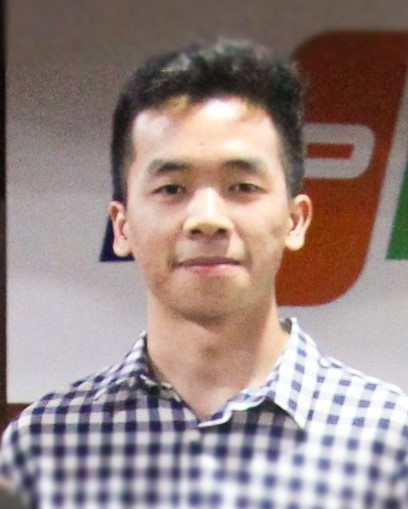}}]{Khiem Le-Huy}
got the Honors Bachelor Degree in Mathematics and Computer Science from the Vietnam National University, Ho Chi Minh City. He was a Research Intern at Smart Health Center, VinBigData JSC, and currently is a Research Assistant at the College of Engineering and Computer Science, VinUniversity, Hanoi, Vietnam. His research interests include Efficient Machine Learning and AI for Biomedical Applications. 
\end{IEEEbiography}

\begin{IEEEbiography}[{\includegraphics[width=1in,height=1.25in,clip,keepaspectratio]{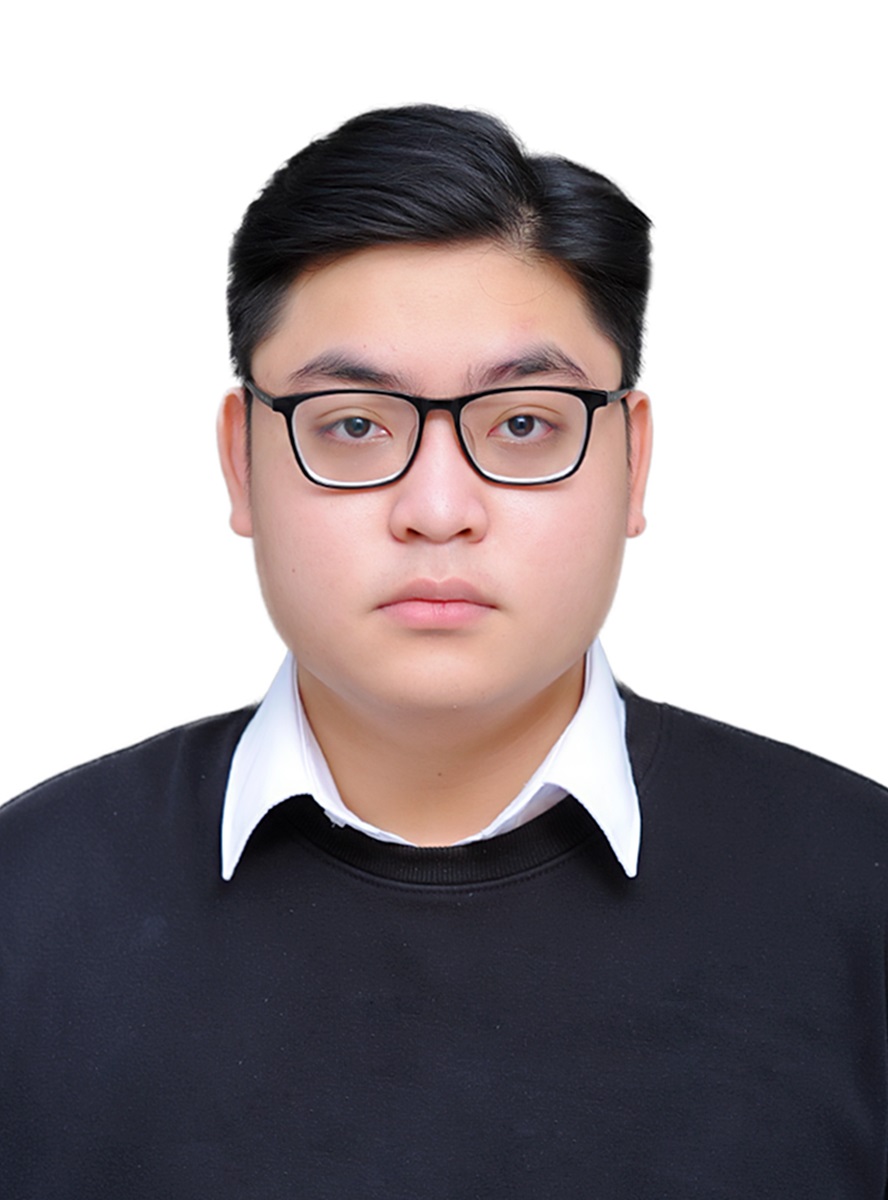}}]{Long Ho-Tuan}
got the Honors Bachelor Degree in Computer Science from the Vietnam National University, Hanoi, Vietnam. Currently, he is a Research Assistant at the College of Engineering and Computer Science, VinUniversity, Hanoi, Vietnam. His research interests include Federated Learning and AI for Biomedical Applications. 
\end{IEEEbiography}

\begin{IEEEbiography}[{\includegraphics[width=1in,height=1.25in,clip,keepaspectratio]{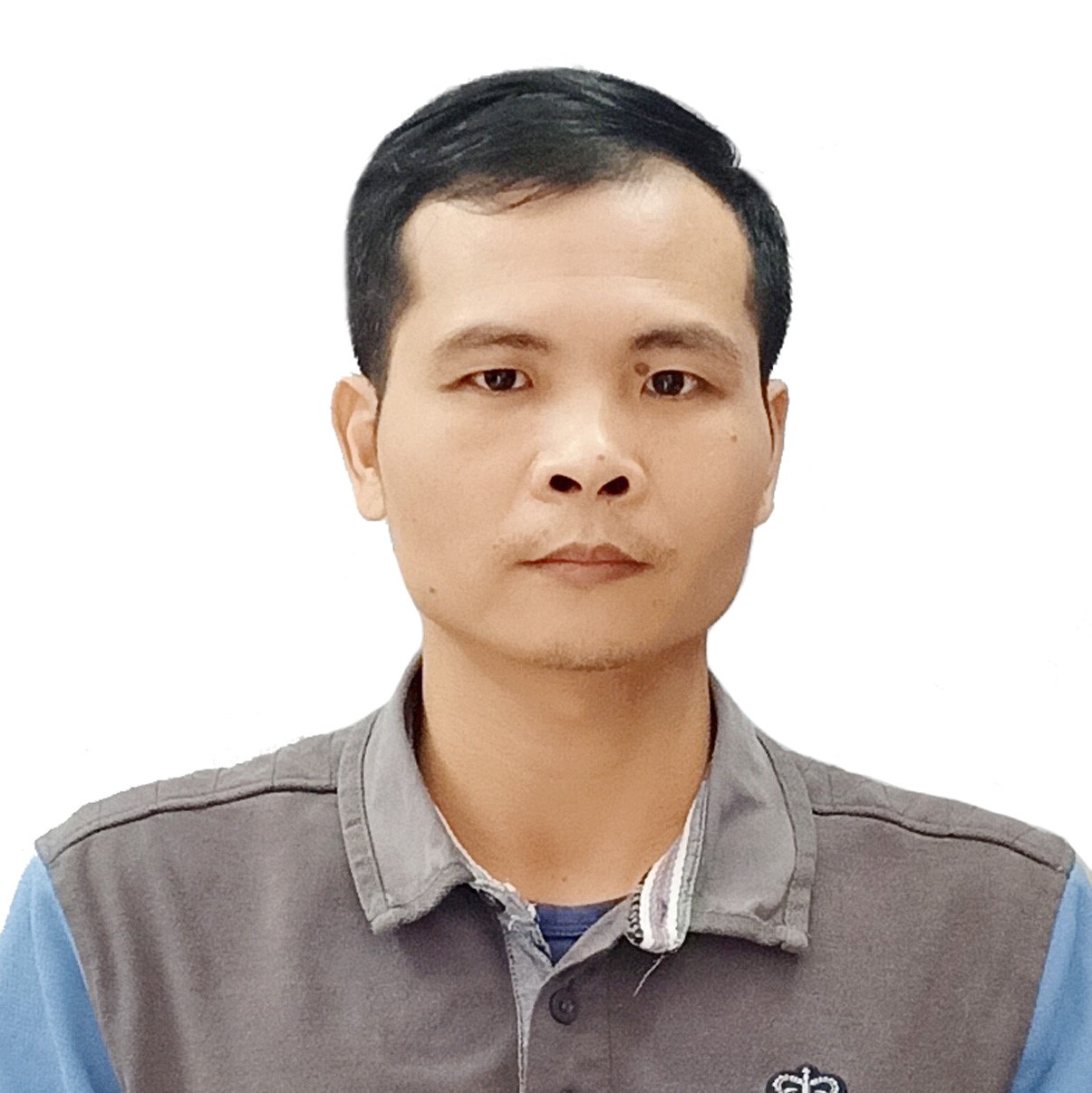}}]{Cuong Do-Danh}(Member, IEEE)
received his B.Sc. degree in electronics and telecommunication from Vietnam National University, Hanoi, Vietnam, in 2004, M.Eng. degree in electronics from Chungbuk National University, Korea, in 2007, and a Ph.D. degree in electronics from Cork Institute of Technology, Ireland, in 2012. He had more than four years working as a postdoc researcher at the University of Cambridge, U.K in both fields of MEMS and CMOS circuits for low-power sensors and timing applications. He is now an Assistant Professor at VinUniversity. His research interests include sensors for medical applications and sensor fusion. 
\end{IEEEbiography}

\begin{IEEEbiography}[{\includegraphics[width=1in,height=1.25in,clip,keepaspectratio]{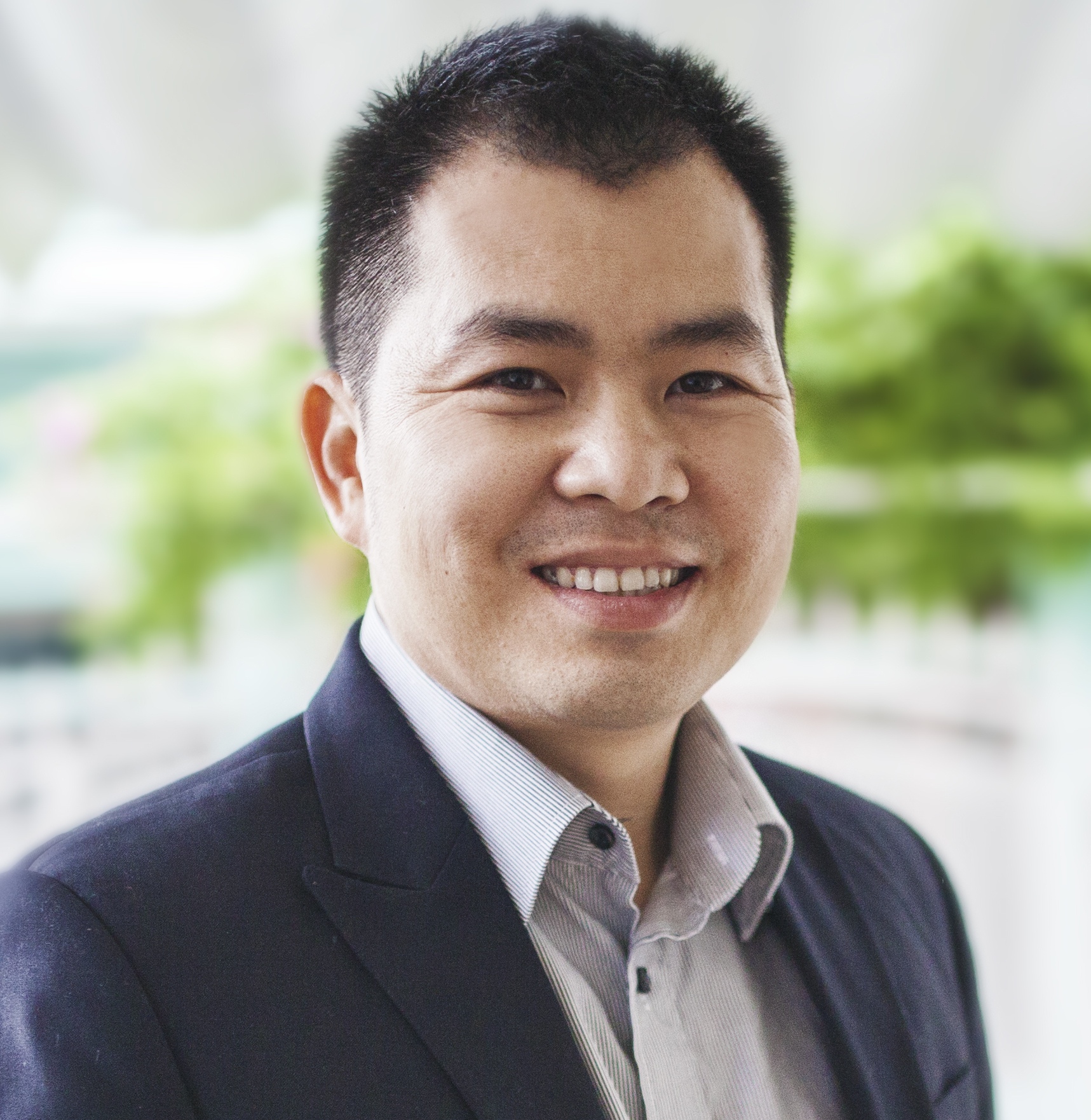}}]{Danh Le-Phuoc}(Member, IEEE) is a DFG Principle Investigator at the Technical University
of Berlin. His research interests include
linked data and the Semantic Computing for IoT-Edge-Cloud continuum, neural-symbolic AI, databases, pervasive
computing, semantic stream processing and reasoning. Le Phuoc received a PhD
in computer science from the National
University of Ireland. Contact him at danh@danhlephuoc.info.
\end{IEEEbiography}

 \onecolumn
\section{Appendix A: Power and Storage}

\begin{figure*}[ht]
    \centering
    \includegraphics[keepaspectratio, width=0.98\textwidth]{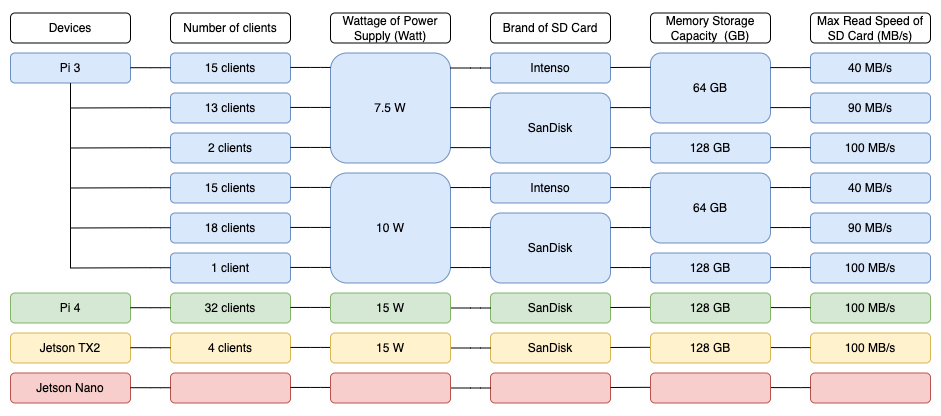}
    \caption{Devices power and storage specification}
    \label{fig:power-storage}
\end{figure*}

The testbed utilized in this research project consists of a diverse array of devices, including Raspberry Pi 3, Raspberry Pi 4, and various models from the NVIDIA Jetson family. These devices are equipped with different types of storage, which come in varying capacities and speeds. Additionally, the devices are powered by a variety of power supplies, with power outputs ranging from 7.5W to 15W. The detailed specification is shown in Figure~\ref{fig:power-storage}

\onecolumn
\section{Appendix B: Compare Convergence Time of Different hardware profile}

\begin{figure}[ht]%
    \centering
    \subfloat[\centering Test Accuracy]{{\includegraphics[keepaspectratio, width=0.3\textwidth]{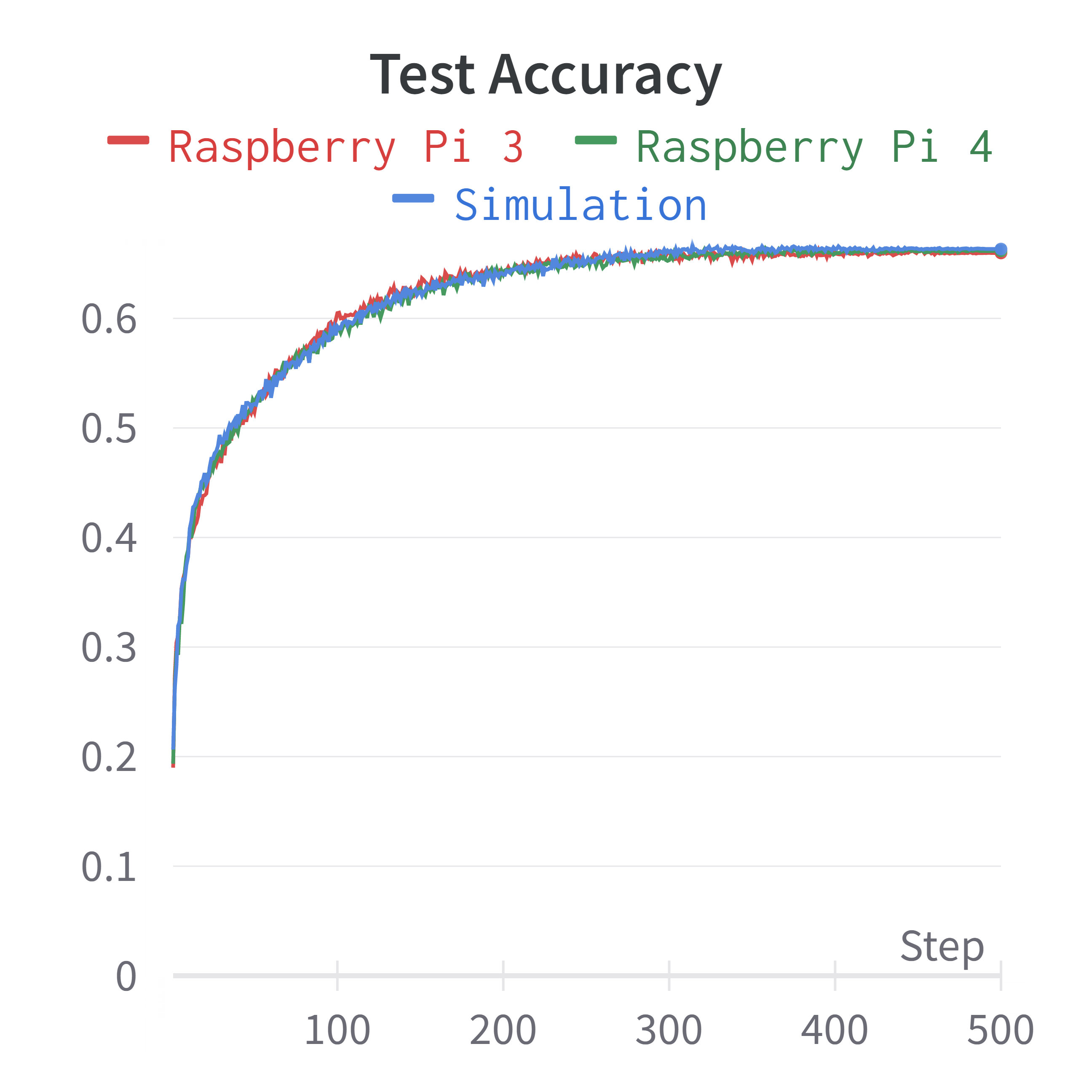} }}%
    \qquad
    \subfloat[\centering Test Loss]{{\includegraphics[keepaspectratio, width=0.3\textwidth]{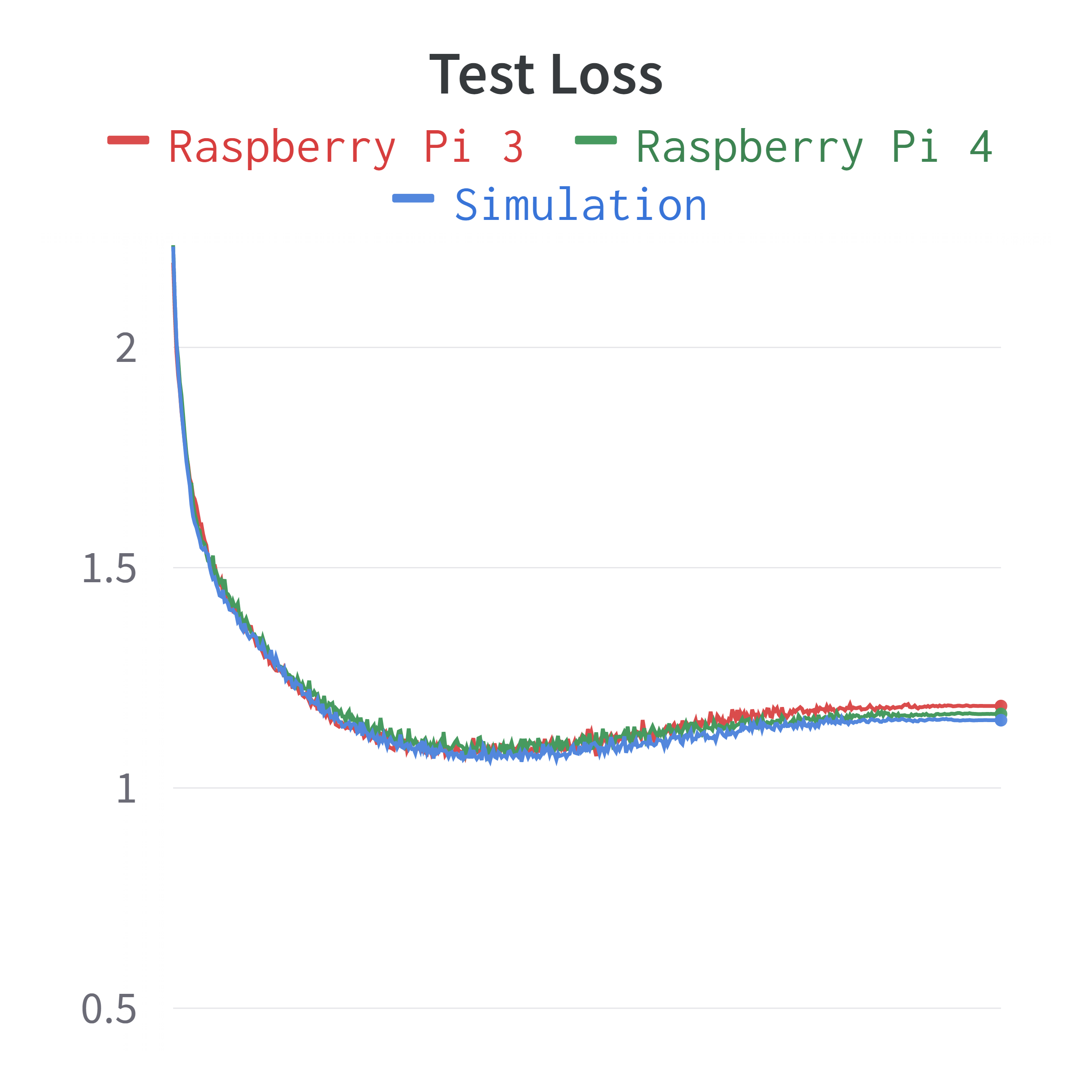} }}%
    \qquad
    \subfloat[\centering Convergence Time]{{\includegraphics[keepaspectratio, width=0.3\textwidth]{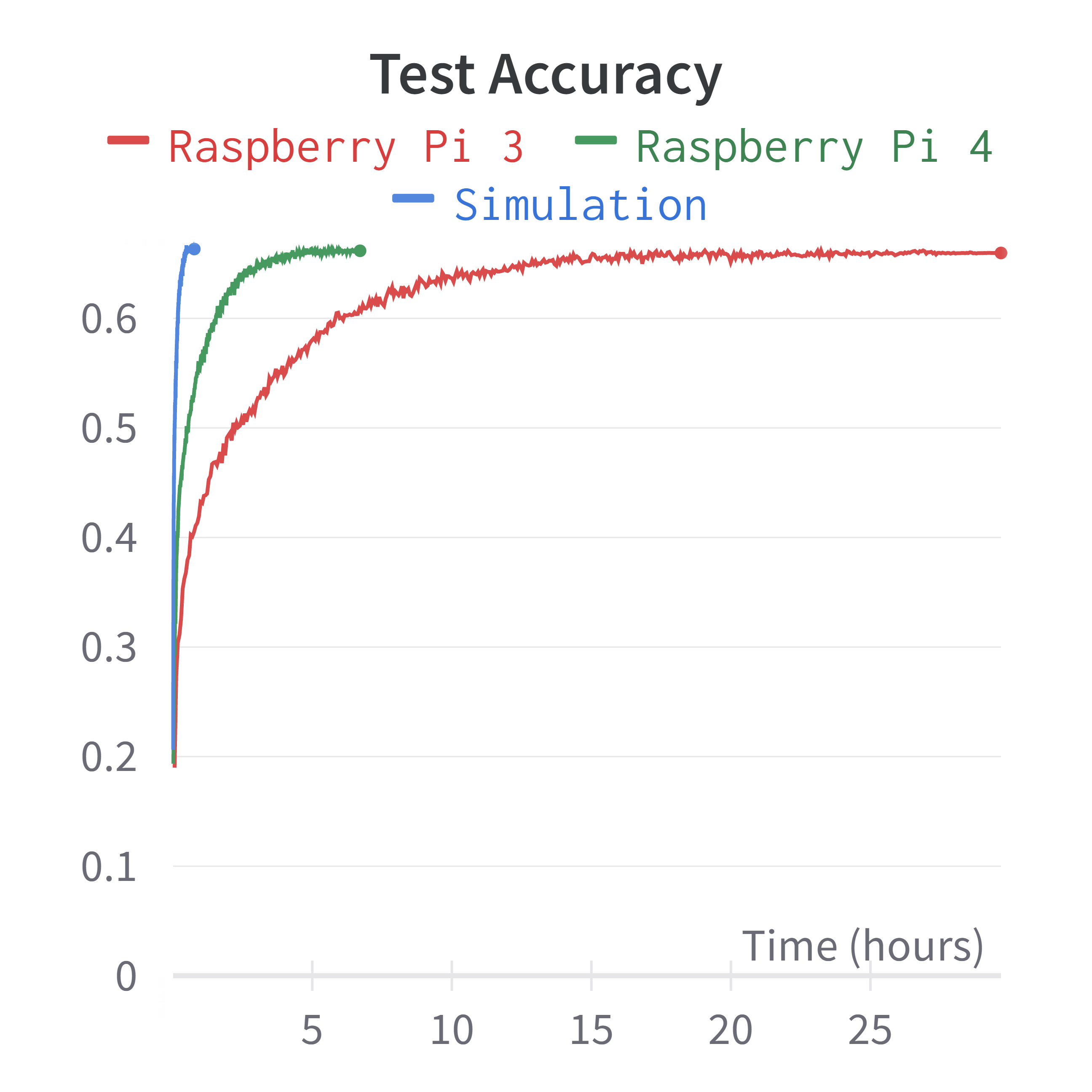} }}%
    \caption{Experiment of Federated Learning on different set of devices. Setup with 8 Raspberry Pi 3 vs 8 Raspberry Pi 4 vs 8 simulation threads of high-end computer on CIFAR10 IID data.}%
    \label{fig:apendixb}%
\end{figure}

This appendix presents the results of an experiment conducted to compare the convergence time of different hardware profiles in Federated Learning. The experiment involved training models on three distinct sets of devices: 8 Raspberry Pi 3 devices, 8 Raspberry Pi 4 devices, and 8 simulation threads on a high-end computer. The experiment used CIFAR10 IID data and monitored the progress of the models on different metrics which is test accuracy, test loss, and convergence speed. Three sub-figures in Figure~\ref{fig:apendixb} provide a visual representation of the performance differences observed among the hardware profiles.

\section{Appendix C: The impact for the number of client}

This appendix presents the findings from an experiment conducted to examine the impact of the number of clients on Federated Learning. The experiment involved training models on Raspberry Pi 3 devices using CIFAR10 IID data, with the number of clients gradually increasing from 8 to 16, 32, and finally 64. 
\begin{figure}[ht]%
    \centering
    \subfloat[\centering Test Accuracy]{{\includegraphics[keepaspectratio, width=0.3\textwidth]{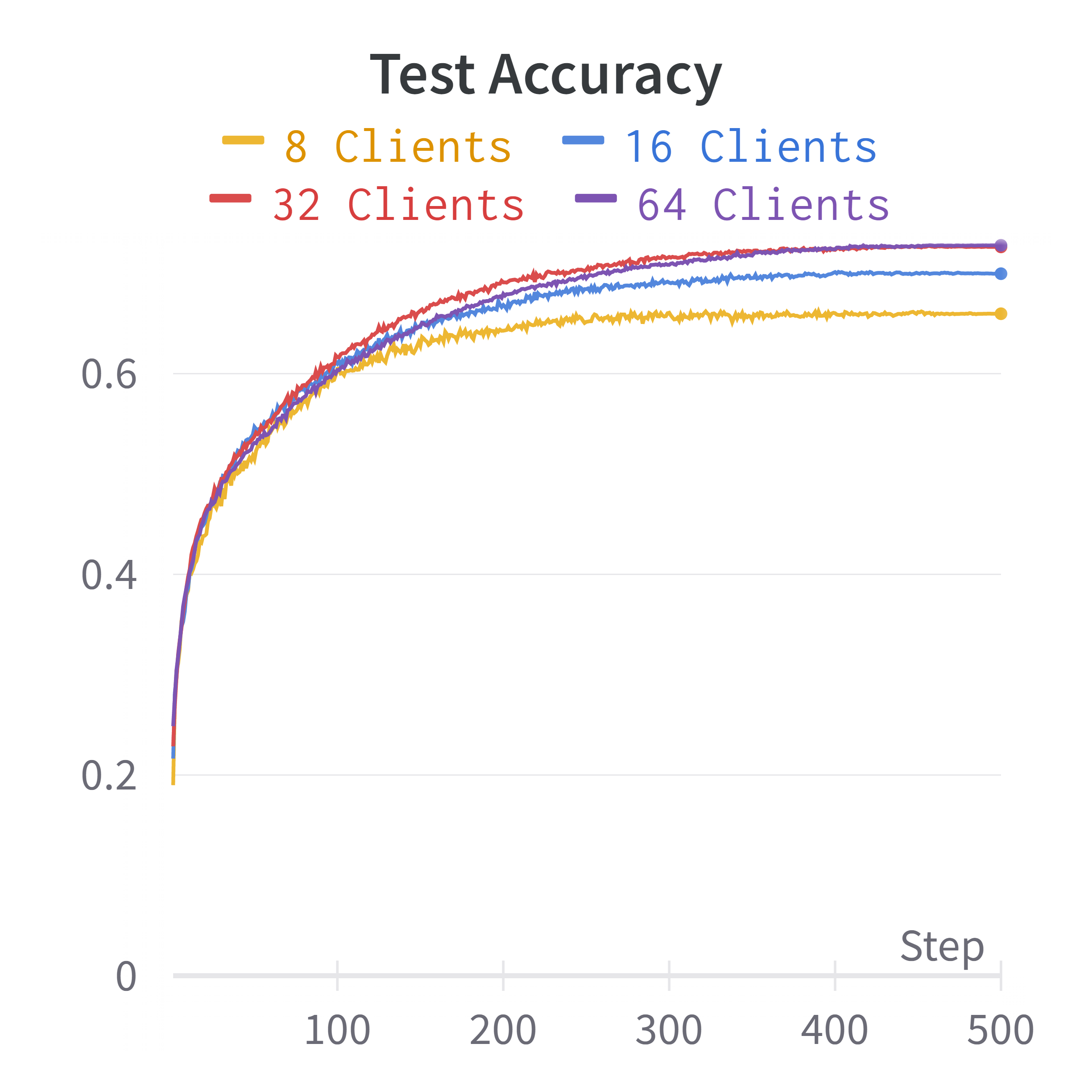} }}%
    \qquad
    \subfloat[\centering Test Loss]{{\includegraphics[keepaspectratio, width=0.3\textwidth]{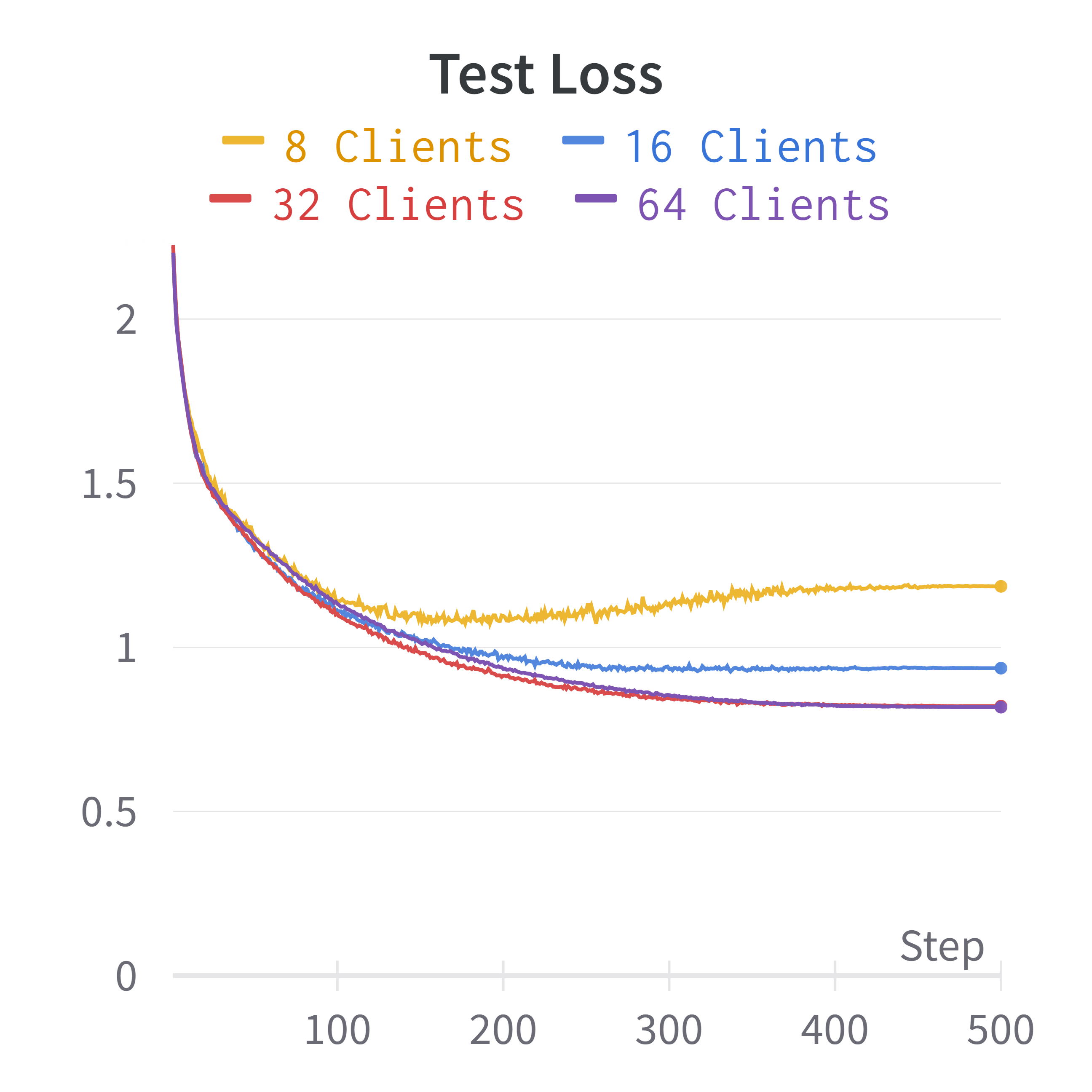} }}%
    \qquad
    \subfloat[\centering Convergence Time]{{\includegraphics[keepaspectratio, width=0.3\textwidth]{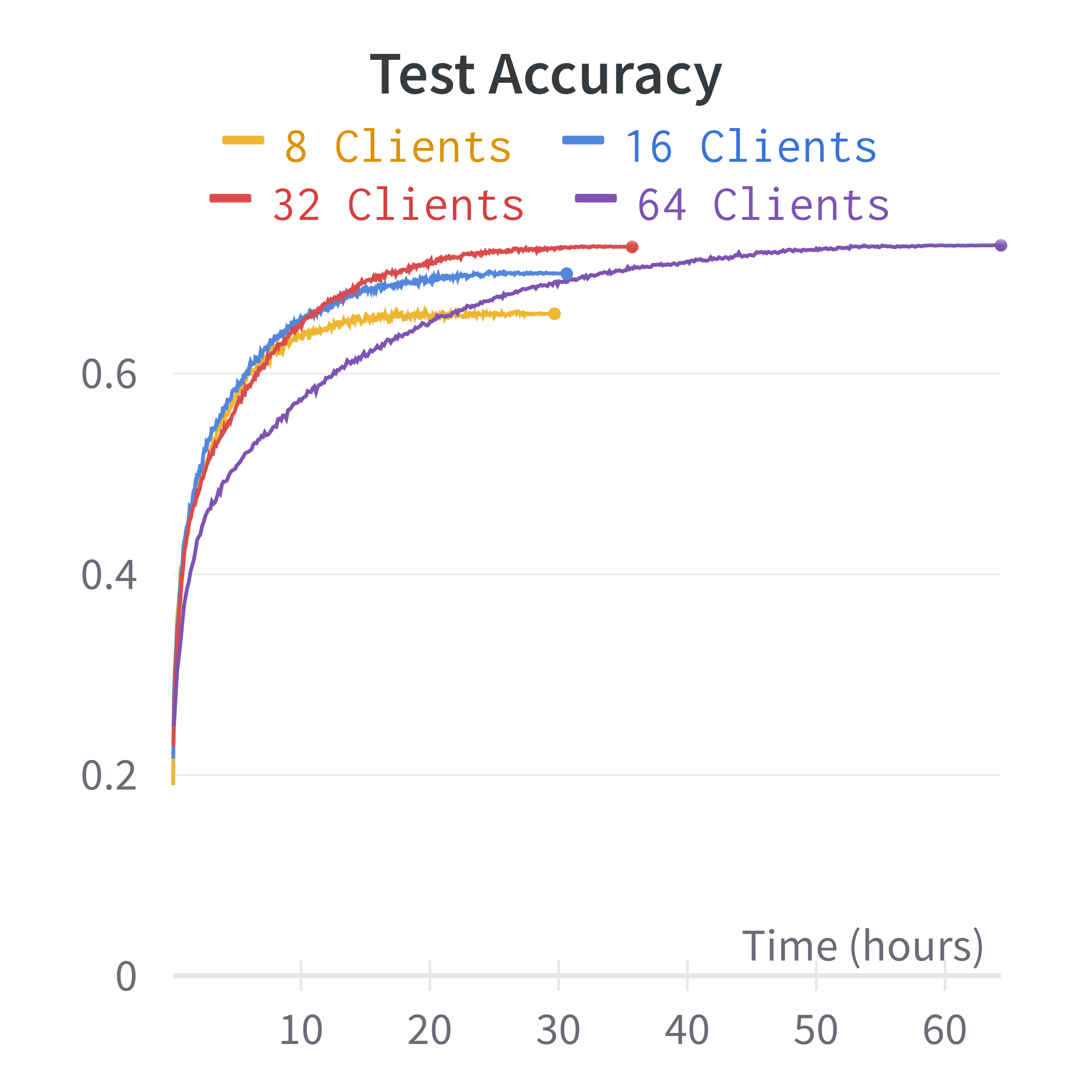} }}%
    \caption{Experiment of Federated Learning on incremental from 8, 16, 32 to 64 clients. Setup on Raspberry Pi 3 using CIFAR10 IID data.}%
    \label{fig:Exps-incremental-device}%
\end{figure}

\section{Appendix D: Heterogeneity in Data and Device profiles}

This appendix examines how differences in both data and device profiles impact Federated Learning. The first figure compares the performance of models trained on different types of data: identical, non-identical, and extremely non-identical. We used 32 Raspberry Pi 4 devices and CIFAR 100 data for this experiment. The figure shows how the models perform under these varying data distributions, highlighting the challenges posed by diverse data.

The second figure explores the performance differences among diverse device profiles. We used a setup with 32 devices, including 32 Raspberry Pi 3, 32 Raspberry Pi 4, and a mix up of 32 devices from different type of devices. The models were trained on CIFAR 10 identical data. This figure compares the performance achieved by these different device profiles, giving insights into the advantages and limitations of diversity in Federated Learning scenarios.

\begin{figure}[ht]%
    \centering
    \subfloat[\centering Test Accuracy]{{\includegraphics[keepaspectratio, width=0.3\textwidth]{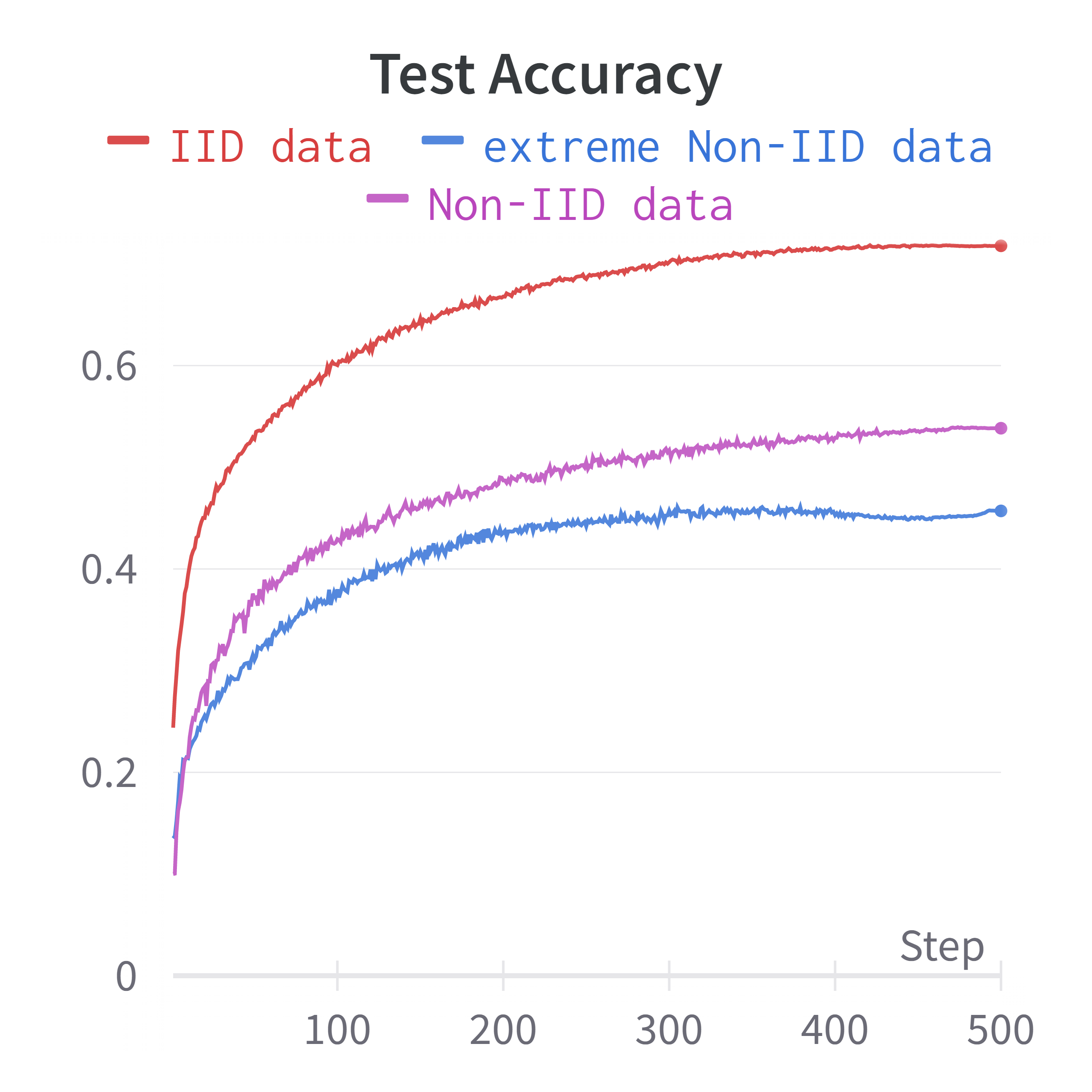} }}%
    \qquad
    \subfloat[\centering Test Loss]{{\includegraphics[keepaspectratio, width=0.3\textwidth]{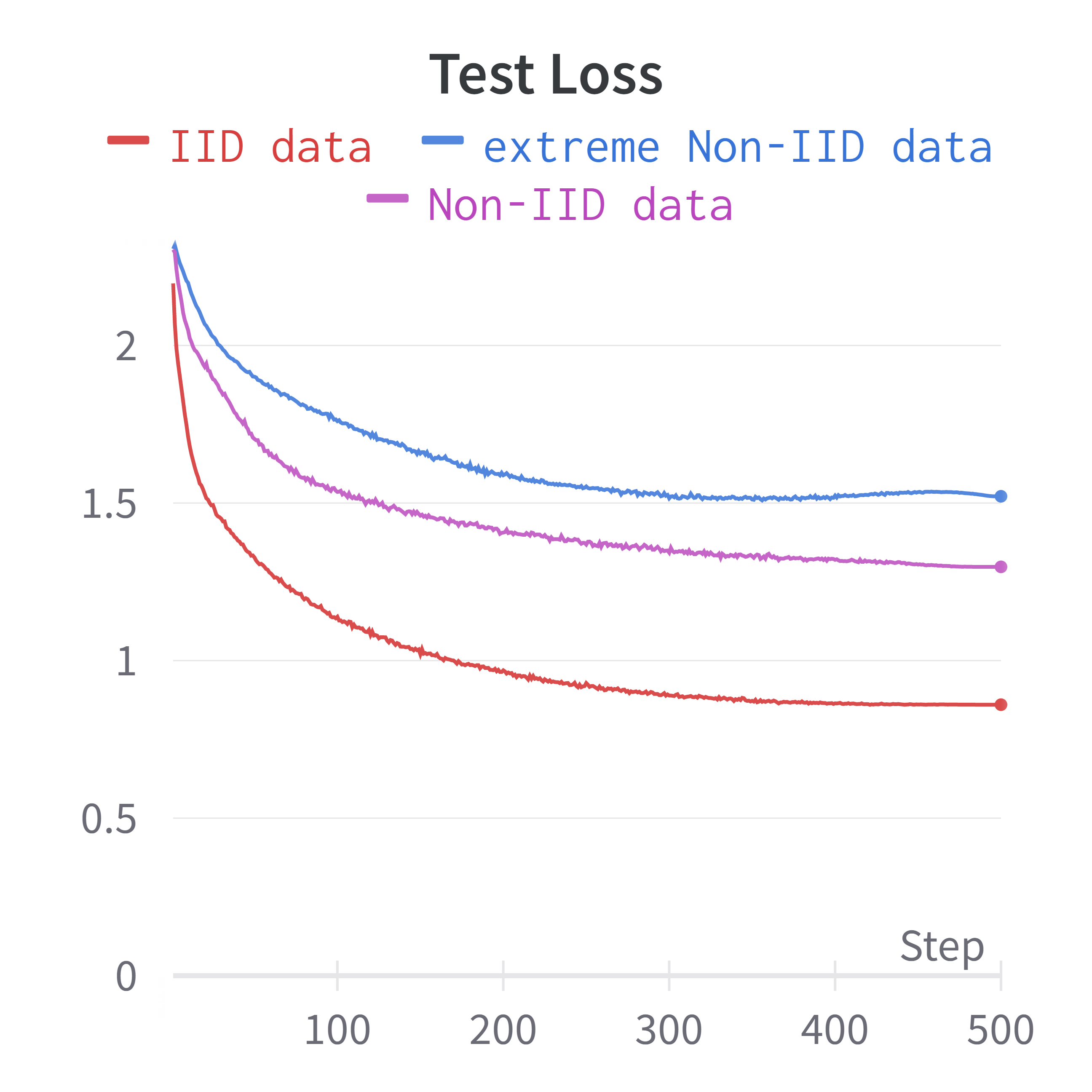} }}%
    \qquad
    \subfloat[\centering Convergence Time]{{\includegraphics[keepaspectratio, width=0.3\textwidth]{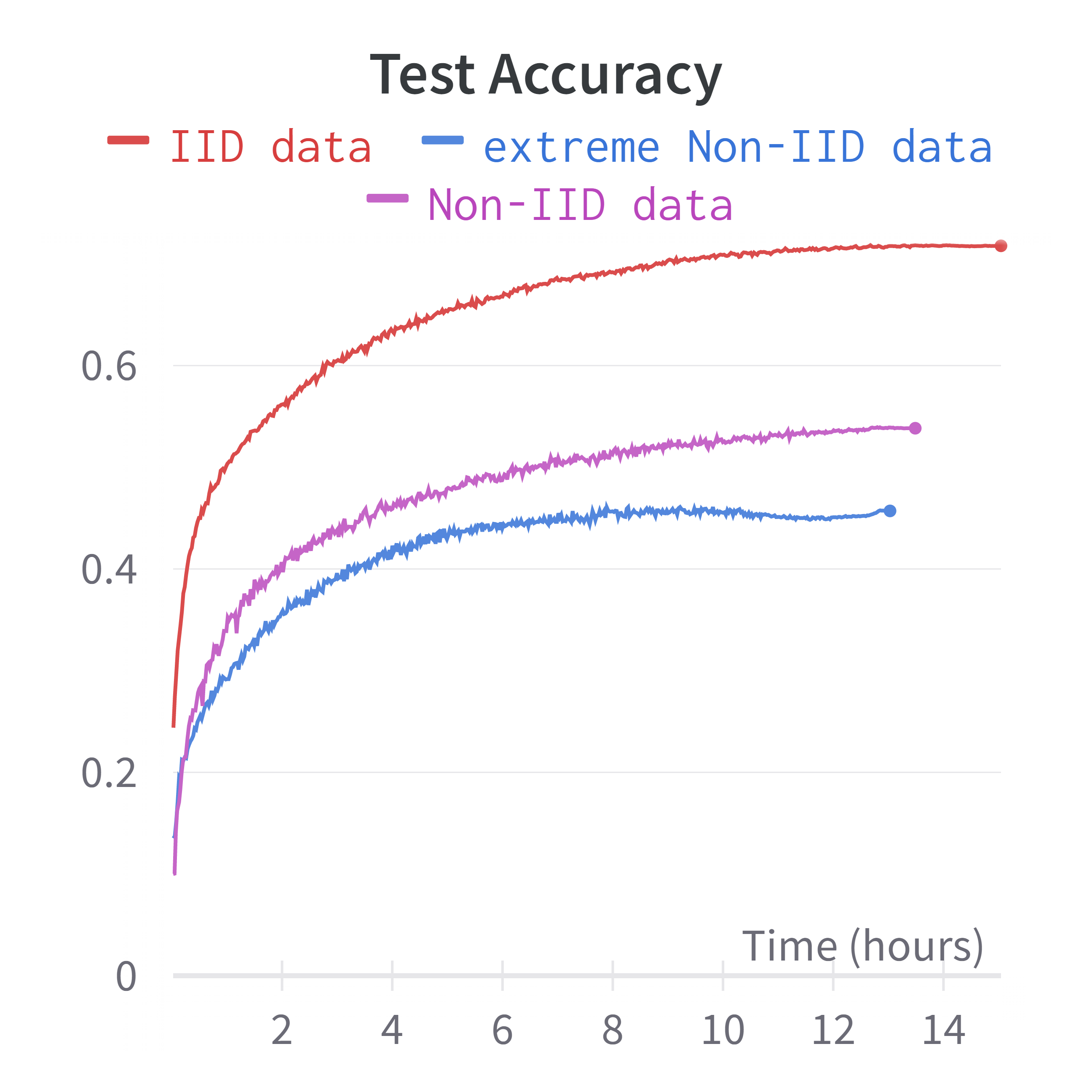} }}%
    \caption{Comparision between IID, Non-IID, and extreme Non-IID data. Setup on 32 Raspberry Pi 4, and CIFAR 100 data.}%
    \label{fig:Exps-diff-device}%
\end{figure}

\begin{figure}%
    \centering
    \subfloat[\centering Test Accuracy]{{\includegraphics[keepaspectratio, width=0.4\textwidth]{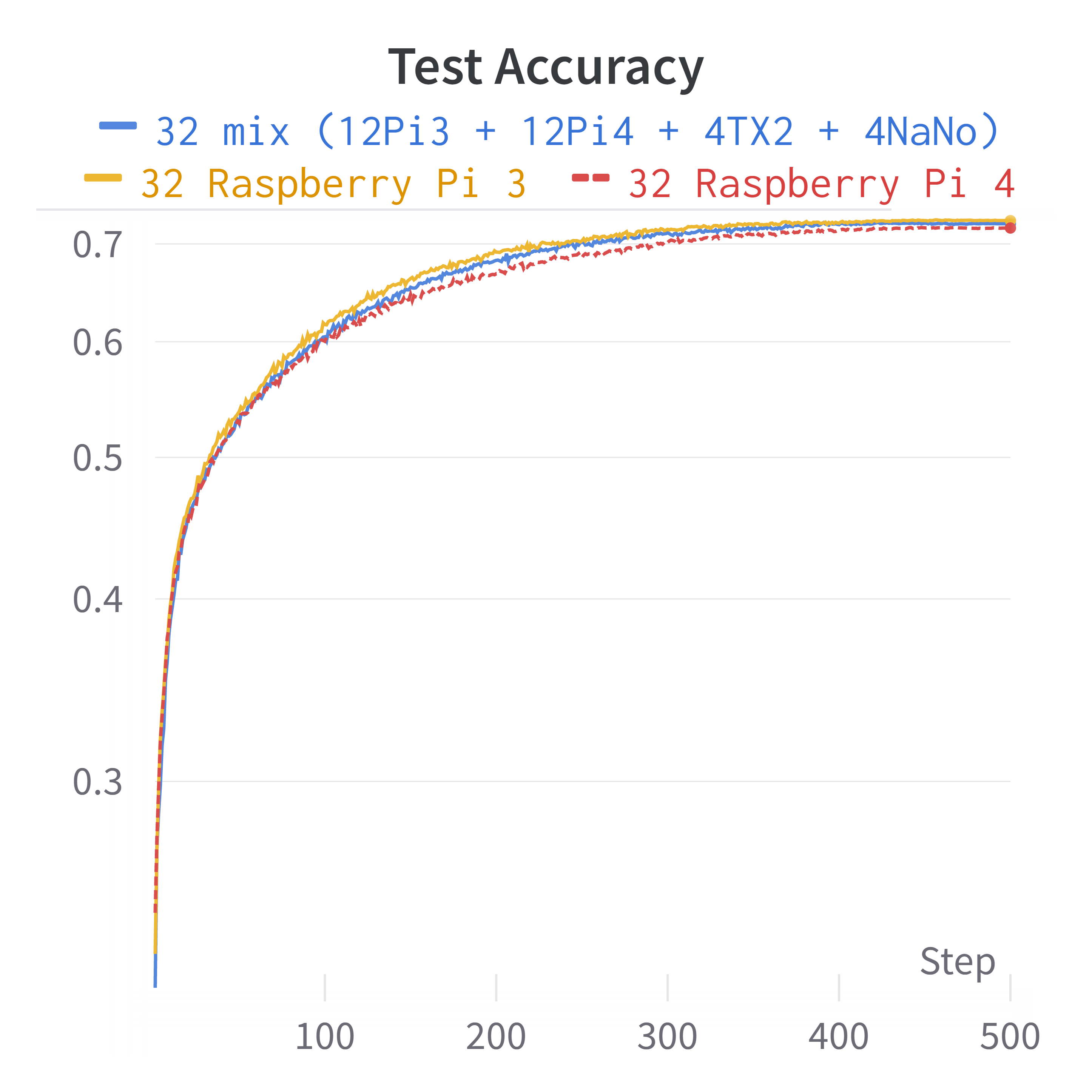} }}%
    \qquad
    \subfloat[\centering Convergence Time]{{\includegraphics[keepaspectratio, width=0.4\textwidth]{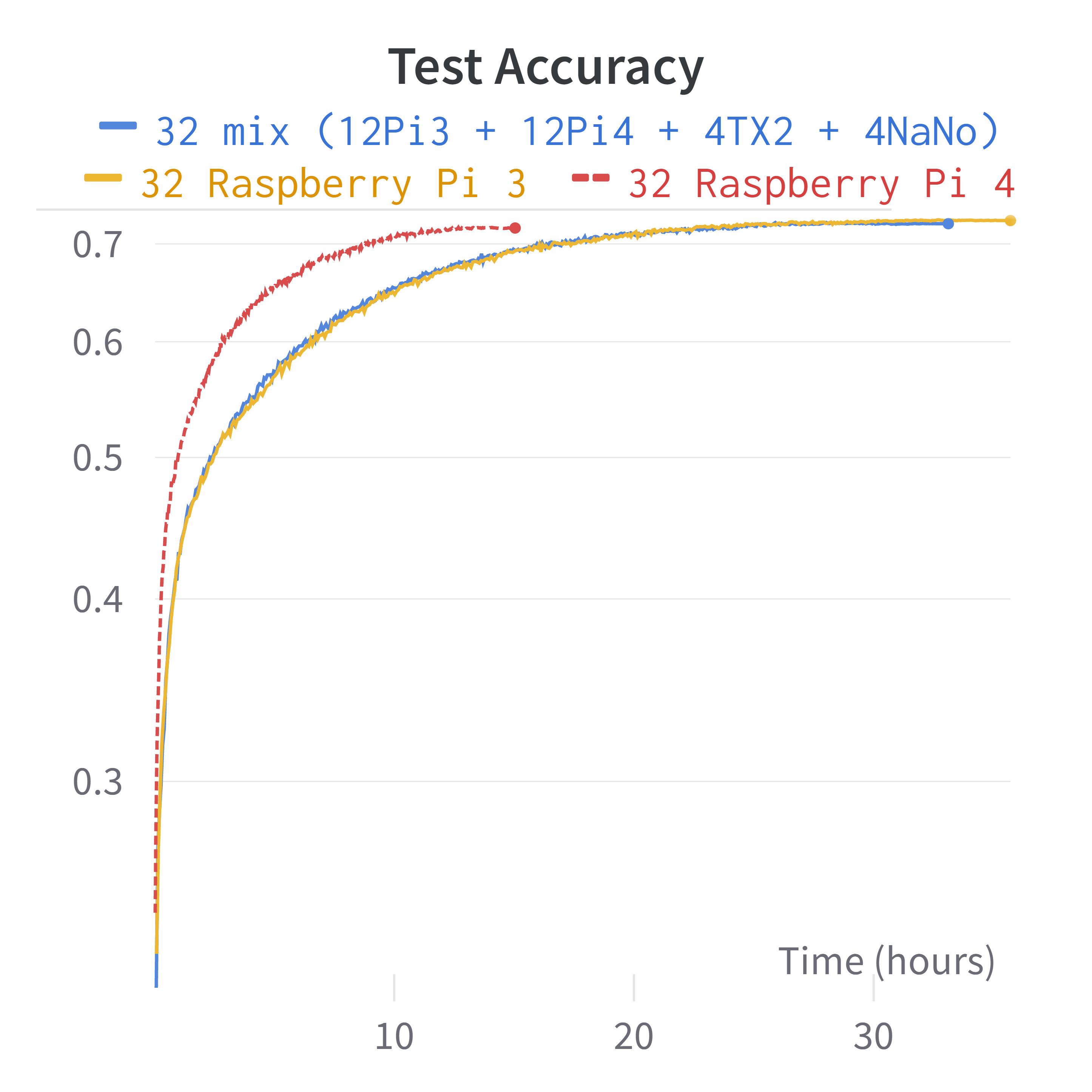} }}%
    \caption{Heterogeneous devices comparison. Setup on 32 Raspberry Pi 3 vs 32 Raspberry Pi 4 vs 32 devices that mix up of 12 Raspberry Pi 3 + 12 Raspberry Pi 4 + 4 NVIDIA Jetson TX2 + 4 NVIDIA Jetson NaNo using CIFAR 10 IID data.}%
    \label{fig:Exps-heterogeneous-devices}%
\end{figure}

\end{document}